\documentclass[journal]{IEEEtran}
\usepackage{cite}
\usepackage[pdftex]{graphicx}
\ifCLASSINFOpdf
\else
\fi
\usepackage{amsmath}
\usepackage{algorithmic}
\usepackage{array}
\usepackage{url}
\usepackage{multirow}

\usepackage{graphicx}

\usepackage{longtable}
\usepackage[colorlinks,
linkcolor=blue, 
anchorcolor=blue, 
citecolor=blue, 
]{hyperref}

\usepackage{threeparttable}
\usepackage{pifont}
\usepackage{amssymb}
\usepackage[numbers,sort&compress]{natbib}
\usepackage{subfigure}
\usepackage{color,xcolor}
\usepackage{colortbl}
\usepackage{booktabs}
\definecolor{il}{HTML}{FD6864}
\definecolor{rl}{HTML}{67FD9A}

\begin{document}
\title{A Survey of Deep RL and IL for Autonomous Driving Policy Learning}
\author{Zeyu Zhu,~\IEEEmembership{Member,~IEEE,}
Huijing Zhao,~\IEEEmembership{Senior Member,~IEEE,}
\thanks{This work is supported in part by the National Natural Science Foundation of China under Grant (61973004).
	The authors are with the Key Laboratory of Machine Perception (MOE) and with the School of Electronics Engineering and Computer Science, Peking University, Beijing 100871, China.
	Correspondence: H.Zhao, zhaohj@pku.edu.cn.}
}

\maketitle

\begin{abstract}
Autonomous driving (AD) agents generate driving policies based on online perception results, which are obtained at multiple levels of abstraction, e.g., behavior planning, motion planning and control. Driving policies are crucial to the realization of safe, efficient and harmonious driving behaviors, where AD agents still face substantial challenges in complex scenarios. Due to their successful application in fields such as robotics and video games, the use of deep reinforcement learning (DRL) and deep imitation learning (DIL) techniques to derive AD policies have witnessed vast research efforts in recent years. This paper is a comprehensive survey of this body of work, which is conducted at three levels: First, a taxonomy of the literature studies is constructed from the system perspective, among which five modes of integration of DRL/DIL models into an AD architecture are identified. Second, the formulations of DRL/DIL models for conducting specified AD tasks are comprehensively reviewed, where various designs on the model state and action spaces and the reinforcement learning rewards are covered. Finally, an in-depth review is conducted on how the critical issues of AD applications regarding driving safety, interaction with other traffic participants and uncertainty of the environment are addressed by the DRL/DIL models. To the best of our knowledge, this is the first survey to focus on AD policy learning using DRL/DIL, which is addressed simultaneously from the system, task-driven and problem-driven perspectives. We share and discuss findings, which may lead to the investigation of various topics in the future.
\end{abstract}

\IEEEpeerreviewmaketitle

\section{Introduction}
\begin{figure}[h]
	\centering
	\includegraphics[width=0.85\linewidth]{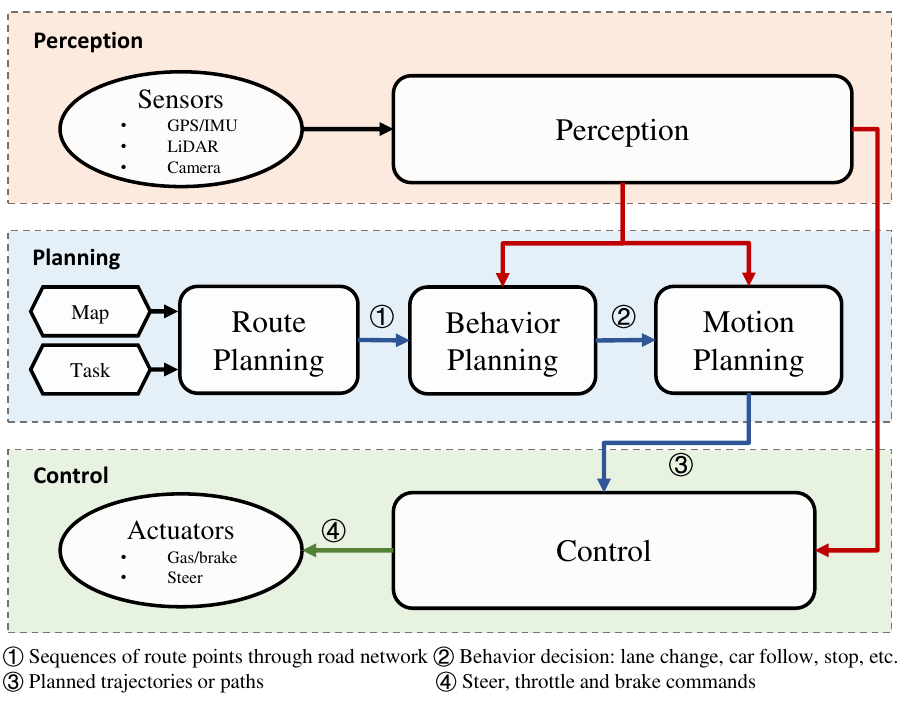}
	\caption{Architecture of autonomous driving systems. A general abstraction that is based on \cite{gonzalez2016review,li2016realtime,paden2016survey,ulbrich2017towards}.}
	\label{traditional_architecture}
	\vspace{-0.5cm}
\end{figure}

\IEEEPARstart{A}{\textbf{utonoumous}} \textbf{driving} (\textbf{AD}) has received extensive attention in recent decades \cite{urmson2008selfdriving,thrun2010towards,eskandarian2012handbook,grigorescu2020survey} and could be a promising solution for improving road safety \cite{world2018global}, traffic flow \cite{talebpour2016influence} and fuel economy \cite{payre2014intention}, among other factors. 
A typical architecture of an AD system is illustrated in Fig. \ref{traditional_architecture}, which is composed of perception, planning and control modules.
An AD agent generates \textbf{driving policies} based on online perception results, which are obtained at multiple levels
of abstraction, e.g., behavior planning, motion planning and control.
The earliest autonomous vehicles can be dated back to \cite{dickmanns1987autonomous,thorpe1988vision,pomerleau1989alvinn}. One milestone was the Defense Advanced Research Projects Agency (DARPA) Grand Challenges \cite{thrun2006stanley,buehler2009darpa}. 
Recent years have witnessed a huge boost in AD research, and many products and prototyping systems have been developed. Despite the fast development of the field, AD still faces substantial challenges in complex scenarios for the realization of safe, efficient and harmonious driving behaviors \cite{li2018humanlike,wilko2018planning}.
\textbf{Reinforcement learning} (\textbf{RL}) is a principled mathematical framework for solving sequential decision making problems \cite{yu1995road,ng2004autonomous,sutton1998reinforcement}. \textbf{Imitation learning} (\textbf{IL}), which is closely related, refers to learning from expert demonstrations. However, the early methods of both were limited to relatively low-dimensional problems. The rise of \textbf{deep learning} (\textbf{DL}) techniques \cite{lecun2015deep,goodfellow2016deep} in recent years has provided powerful solutions to this problem through the appealing properties of \textbf{deep neural networks} (\textbf{DNNs}): function approximation and representation learning. DL techniques enable the scaling of RL/IL to previously intractable problems (e.g., high-dimensional state spaces), which have increased in popularity for complex locomotion \cite{lillicrap2016continuous}, robotics \cite{zhu2017target} and autonomous driving \cite{liang2018cirl,chen2019attention,kendall2019learning} tasks. Unless otherwise stated, this survey focuses on \textbf{Deep RL} (\textbf{DRL}) and \textbf{Deep IL} (\textbf{DIL}).

A large variety of DRL/DIL models have been developed for learning AD policies, which are reviewed in this paper. Several surveys are relevant to this study. \cite{gonzalez2016review,paden2016survey} survey the motion planning and control methods of automated vehicles before the era of DL. \cite{arulkumaran2017deep,li2017deep,nguyen2018deep,hussein2017imitation,osa2018algorithmic} review general DRL/DIL methods without considering any particular applications. \cite{grigorescu2020survey} addresses the deep learning techniques for AD with a focus on perception and control, while \cite{kuutti2019survey} addresses control only. \cite{kiran2020deep} provides a taxonomy of AD tasks to which DRL models have been applied and highlights the key challenges. However, none of these studies answers the following questions:

\textit{How can DRL/DIL models be integrated into AD systems from the perspective of system architecture?
How can they be formulated to accomplish specified AD tasks? 
How can methods be designed that address the challenging issues of AD, such as safety, interaction with other traffic participants, and uncertainty of the environment?}
\begin{figure*}[ht]
\centering
\includegraphics[width=\linewidth]{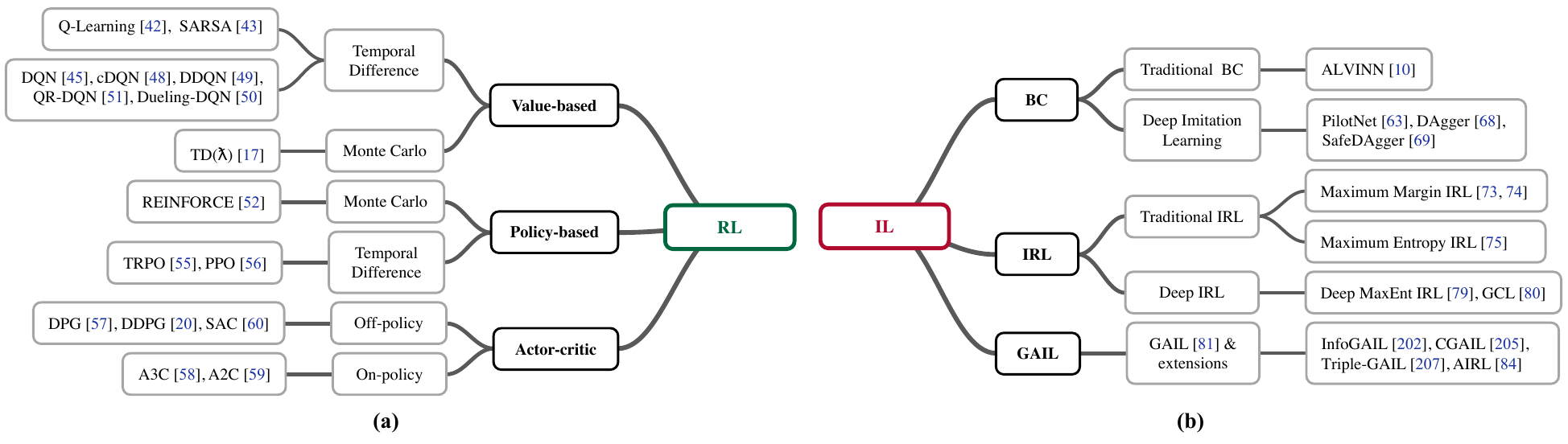}
\caption{A taxonomy of the general methods of reinforcement learning (RL) and imitation learning (IL)}
\label{fig:review_of_rl_il}
\vspace{-0.6cm}
\end{figure*}

This study seeks answers to the above questions. To the best of our knowledge, this is the first survey to focus on AD policy learning using DRL/DIL, which is addressed from the system, task-driven and problem-driven perspectives.
Our contributions are threefold:

\begin{itemize}
\item A taxonomy of the literature is presented from the system perspective, from which five modes of integration of DRL/DIL models into an AD architecture are identified. The studies on each mode are reviewed, and the architectures are compared. It is found that the vast research efforts have focused mainly on exploring the potential of DRL/DIL in accomplishing AD tasks, while intensive studies are needed on the optimization of the architectures of DRL/DIL embedded systems toward real-world applications.
\item The formulations of DRL/DIL models for accomplishing specified AD tasks are comprehensively reviewed, where various designs on the model state and action spaces and the reinforcement learning rewards are covered. It is found that these formulations rely heavily on empirical designs, which are brute-force approaches and lack theoretic support. Changing the designs or tuning the parameters could result in substantially different driving policies, which may pose large challenges to the AD system’s stability and robustness in real-world deployment.
\item The critical issues of AD applications that are addressed by the DRL/DIL models regarding driving safety, interaction with other traffic participants and uncertainty of the environment are comprehensively discussed. It is found that driving safety has been well studied. A typical strategy is to combine with traditional methods to ensure a DRL/DIL agent’s functional safety; however, striking a balance between optimal policies and hard constraints remains non-trivial. The studies on the latter two issues are still highly preliminary, in which the problems have been addressed from divergent perspectives and the studies have not been conducted systematically. 
\end{itemize}

The remainder of this paper is organized as follows: Sections \ref{sec:review of rl} and \ref{sec:review of il} briefly introduce (D)RL and (D)IL, respectively. Section \ref{sec:architecture} reviews the research on DRL/DIL in AD from a system architecture perspective. Section \ref{sec: task-driven} reviews the task-driven methods and examines the formulations of DRL/DIL models for completing specified AD tasks. Section \ref{sec:problem-driven} reviews problem-driven methods, in which specified autonomous vehicle problems and challenges are addressed. Section \ref{sec:challenge} discusses the remaining challenges, and the conclusions of the survey are presented in Section \ref{sec:conclusion}. 
\section{Preliminaries of Reinforcement Learning}
\label{sec:review of rl}
\subsection{Problem Formulation}
Reinforcement learning (RL) is a principled mathematical framework that is based upon the paradigm of trial-and-error learning, where an agent interacts with its environment through a trade-off between exploitation and exploration \cite{thrun1992efficient,coggan2004exploration,zhang18diversity}.
\textbf{Markov decision processes (MDPs)} are a mathematically idealized form of RL \cite{sutton1998reinforcement}, which are represented as $(\cal S, \cal A, \cal P, \cal R, \gamma)$, where $\cal S$ and $\cal A$ denote the sets of states and actions, respectively, and $\cal P$$(s_{t+1}|s_t,a_t):\cal S \times \cal S \times \cal A \rightarrow$ $[0,1]$ is the \textbf{transition/dynamics function} that maps state-action pairs onto a distribution of next-step states. The numerical immediate reward function $\cal R$$(s_t,a_t,s_{t+1}):\cal S \times \cal A \times \cal S\rightarrow \mathbb{R}$ serves as a learning signal. A discount factor $\gamma \in [0,1]$ determines the present value of future rewards (lower values encourage more myopic behaviors).
MDPs' states satisfy the \textbf{Markov property} \cite{sutton1998reinforcement}, namely, future states depend only on the immediately preceding states and actions.
\textbf{Partially observable MDPs (POMDPs)} extend MDPs to problems in which access to fully observable Markov property states is not available.
A POMDP has an observation set $\Omega$ and an observation function $\cal O$, where $\cal O$$(a_t,s_{t+1},o_{t+1})=p(o_{t+1}|a_t,s_{t+1})$ is the probability of observing $o_{t+1}$ given that the agent has executed action $a_t$ and reached state $s_{t+1}$ \cite{shani2013survey}.
For the theory and algorithms of POMDPs, we refer readers to  \cite{lovejoy1991survey,shani2013survey}.

The MDP agent selects an action $a_t \in \cal A$ at each time step $t$ based on the current state $s_t$, receives a numerical reward $r_{t+1}$ and visits a new state $s_{t+1}$. The generated sequence $\{s_0,a_0,r_1,s_1,a_1,r_2,...\}$ is called a \textbf{rollout} or \textbf{trajectory}.
The expected cumulative reward in the future, namely, the \textbf{expected discounted return} $G_t$ after time step $t$, is defined as \cite{sutton1998reinforcement}:
\begin{eqnarray}
G_t \doteq r_{t+1} + \gamma r_{t+2} + \gamma^2 r_{t+3} + ... = \sum_{k=0}^{T}\gamma^kr_{t+k+1}
\end{eqnarray}
where $T$ is a finite value or $\infty$ for finite and infinite horizon problems, respectively. The \textbf{policy} $\pi(a|s)$ maps states to probabilities of selecting each possible action. The \textbf{value function} $v_\pi(s)$ is the expected return following $\pi$ from state $s$:
\begin{eqnarray}
\label{eqn_v_return}
v_\pi(s)\doteq \mathbb 
{E}_\pi[G_t|s_t=s]
\end{eqnarray}
Similarly, the \textbf{action-value function} $q_\pi(s,a)$ is defined as:
\begin{equation}
\label {eqn_q_values}
q_\pi(s,a) \doteq \mathbb{E}_\pi[G_t|s_t=s,a_t=a]
\end{equation}
which satisfies the recursive \textbf{Bellman equation} \cite{bellman1957on} :
\begin{equation}
\label{bellman_q}
q_\pi(s_t,a_t)=\mathbb {E}_{s_{t+1}}[r_{t+1} + \gamma q_\pi(s_{t+1},\pi(s_{t+1}))]
\end{equation}
The objective of RL is to identify the optimal policy that maximizes the expected return $\pi^* = \arg \max_\pi \mathbb {E}_\pi[G_t|s_t=s]$. The methods can be divided into three groups, as shown in Fig. \ref{fig:review_of_rl_il} (a).

\subsection{Value-based Methods}

To solve a reinforcement learning problem, one can identify an optimal action-value function and recover the optimal policy from the learned state-action values.
\begin{eqnarray}
\label{value-based-objective}
q_{\pi^*}(s,a) = \max_\pi q_\pi(s,a)\\
{\pi^*}(s) = \arg \max_a q_{\pi^*}(s,a)
\end{eqnarray}
Q-learning \cite{watkins1992technical} is one of the most popular methods, which estimates Q values through temporal difference (\textbf{TD}):
\begin{eqnarray}
\label{eqn_q_update}
q_\pi(s_t,a_t) &\leftarrow& q_\pi(s_t,a_t) + \alpha (Y - q_\pi(s_t,a_t))
\end{eqnarray}
where $\displaystyle Y = r_{t+1} + \gamma \max\limits_{a_{t+1} \in \cal A}q_\pi(s_{t+1},a_{t+1})$ is the temporal difference target and $\alpha$ is the learning rate. This can be regarded as a standard regression problem in which the error to be minimized is $Y - q_\pi(s_t,a_t)$. Q-learning is \textbf{off-policy} since it updates $q_\pi$ based on experiences that are not necessarily generated by the derived policy, while SARSA \cite{rummery1994line} is an \textbf{on-policy} algorithm that uses the derived policy to generate experiences. Another distinction is that SARSA uses target $Y = r_{t+1} + \gamma q_\pi(s_{t+1},a_{t+1})$.
In contrast to TD methods, Monte Carlo methods estimate the expected return through averaging the results of multiple rollouts, which can be applied to non-Markovian episodic environments. TD and Monte Carlo have been further combined in TD($\lambda$) \cite{sutton1998reinforcement}. 

The early methods \cite{watkins1992technical,rummery1994line,sutton1998reinforcement} of this group rely on tabular representations. A major problem is the ``curse of dimensionality" \cite{bellman1966dynamic}, namely, an increase in the number of state features would result in exponential growth of the number of state-action pairs that must be stored. Recent methods use DNNs to approximate a parameterized value function $q (s,a;\omega)$, of which Deep Q-networks (DQNs) \cite{mnih2015human} are the most representative, which learn the values by minimizing the following loss function:
\begin{eqnarray}
\label{value-based-learning}
\displaystyle J(\omega) &=& \mathbb E_t[(Y-q(s_t,a_t;\omega))^2] 
\end{eqnarray}
where $Y=r_{t+1}+\gamma \max q(s_{t+1},a_{t+1};\omega^-)$ is the target, $\omega^-$ denotes the parameters of the target network, and $\omega$ can be learnt based on the gradients.
\begin{equation}
\label {eqn_value_gradient}
\displaystyle \omega \leftarrow \omega - \alpha \mathbb E_t[(Y-q(s_t,a_t;\omega))\nabla q(s_t,a_t;\omega)]
\end{equation}
The major contributions of DQN are the introduction of the target network and experience replay.
To avoid rapidly fluctuating Q values and stabilize the training, the target network is fixed for a specified number of iterations during the update of the primary Q-network and subsequently updated to match the primary Q-network. Moreover, experience replay \cite{lin1992self}, which maintains a memory that stores transitions $(s_t,a_t,s_{t+1},r_{t+1})$, increases the sample efficiency. A later study improves the uniform sample experience replay by introducing priority \cite{schaul2015prioritized}. Continuous DQN (cDQN) \cite{gu2016continuous} derives a continuous variant of DQN based on normalized advantage functions (NAFs). Double DQN (DDQN) \cite{hasselt2016deep} addresses the overestimation problem of DQN through the use of a double estimator. Dueling-DQN \cite{wang2016dueling} introduces a dueling architecture in which both the value function and associated advantage function are estimated. QR-DQN \cite{dabney2018distributional} utilizes distributional reinforcement learning to learn the full value distribution rather than only the expectation.
\subsection{Policy-based Methods}
Alternatively, one can directly search and optimize a parameterized policy $\pi_{\theta}$ to maximize the expected return:
\begin{equation}
\label{policy-based-objective}
\max_\theta J(\theta) = \max_\theta v_{\pi_\theta}(s) =\max_\theta \mathbb E_{\pi_\theta}[G_t|s_t=s]
\end{equation}
where $\theta$ denotes the policy parameters, which can be optimized based on the policy gradient theorem \cite{sutton1998reinforcement}:
\begin{eqnarray}
\label{policy-based-learning}
\displaystyle \nabla J(\theta) &\propto& \sum_s \mu(s) \sum_a q_\pi(s,a)\nabla \pi_\theta(a|s) \nonumber \\
\displaystyle &=& \mathbb E_\pi[\sum_a q_\pi(s_t,a)\nabla \pi_\theta(a|s_t)] \nonumber \\
\displaystyle &=& \mathbb E_\pi[G_t \nabla \ln\pi_\theta(a_t|s_t)]
\end{eqnarray}
where $\mu(s)$ denotes the state visitation frequency.
REINFORCE \cite{williams1992simple} is a straightforward Monte Carlo policy-based method that selects the discounted return $G_t$ following the policy $\pi_\theta$ to estimate the policy gradient in Eqn.\ref{policy-based-learning}. The parameters are updated as follows \cite{sutton1998reinforcement}:
\begin{equation}
\label {eqn_policy_gradient}
\theta \leftarrow \theta + \alpha G_t \nabla \ln\pi_\theta(a_t|s_t)
\end{equation}
This update intuitively increases the log probability of actions that lead to higher returns.
Since empirical returns are used, the resulting gradients suffer from high variances. A common technique for reducing the variance and accelerating the learning is to replace $G_t$ in Eqn. \ref{policy-based-learning} and \ref{eqn_policy_gradient} by $G_t - b(s_t)$ \cite{williams1992simple, sutton1998reinforcement}, where $b(s_t)$ is a baseline. Alternatively, $G_t$ can be replaced by the advantage function \cite{baird1994reinforcement,schulman2016high} $ A_{\pi_\theta}(s,a)=q_{\pi_\theta(s,a)} - v_{\pi_\theta}(s)$.

One problem of policy-based methods is poor gradient updates may result in newly updated policies that deviate wildly from previous policies, which may decrease the policy performance. Trust region policy optimization (TRPO) \cite{schulman2015trust} solves this problem through optimization of a surrogate objective function, which guarantees the monotonic improvement of policy performance. Each policy gradient update is constrained by using a quadratic approximation of the Kullback-Leibler (KL) divergence between policies. Proximal policy optimization (PPO) \cite{schulman2017proximal} improved upon TRPO through the application of an adaptive penalty on the KL divergence and a heuristic clipped surrogate objective function. The requirement for only a first-order gradient also renders PPO easier to implement than TRPO.
\subsection {Actor-Critic Methods}
Actor-critic methods have the advantages of both value-based and policy-based methods, where a neural network \textbf{actor} $\pi_\theta(a|s)$ selects actions and a neural network \textbf{critic} $q(s,a;\omega)$ or $v(s;\omega)$ estimates the values.
The actor and critic are typically updated alternately according to Eqn. \ref{policy-based-learning} and Eqn. \ref{value-based-learning}, respectively.
Deterministic policy gradient (DPG) \cite{silver2014deterministic} is an off-policy actor-critic algorithm that derives deterministic policies. Compared with stochastic policies, DPG only integrates over the state space and requires fewer samples in problems with large action spaces. Deep deterministic policy gradient (DDPG) \cite{lillicrap2016continuous} utilizes DNNs to operate on high-dimensional state spaces with experience replay and a separate actor-critic target network, which is similar to DQN.
Exploitation of parallel computation is an alternative to experience replay. Asynchronous advantage actor-critic (A3C) \cite{mnih2016asynchronous} uses advantage estimates rather than discounted returns in the actor-critic framework and asynchronously updates policy 
and value networks
on multiple parallel threads of the environment.

The parallel independent environments stabilize the learning process and enable more exploration. Advantage actor critic (A2C) \cite{wang2017learning}, which is the synchronous version of A3C, uses a single agent for simplicity or waits for each agent to finish its experience to collect multiple trajectories. Soft actor critic (SAC) \cite{haarnoja2018soft} benefits from the addition of an entropy term to the reward function to encourage better exploration.

\section{Preliminaries of Imitation Learning}
\label{sec:review of il}
\subsection{Problem Formulation}
Imitation learning possesses a simpler form and is based on learning from demonstrations (LfD) \cite{argall2009survey}. 
It is attractive for AD applications, where interaction with the real environment could be dangerous and vast amount of human driving data can be easily collected \cite{john2006NGSIM}. 
A demonstration dataset ${\cal D}=\{\xi_i\}_{i=0..N}$ contains a set of trajectories, where each trajectory $\xi_i=\{(s_t^i,a_t^i)\}_{t=0..T}$ is a sequence of state-action pairs, 
and action $a_t^i \in \cal A$ is taken by expert at state $s_t^i \in S\cal$ under the guidance of an underlying policy $\pi_E$ of the expert \cite{hussein2017imitation}.
Using the collected dataset $\cal D$, a common optimization-based strategy of imitation learning is to learn a policy $\pi^*: \cal S \rightarrow \cal A$ that mimics the expert's policy $\pi_E$ by satisfying \cite{osa2018algorithmic}
\begin{equation}
\pi^* = \arg \min_\pi \mathbb D(\pi_E, \pi)
\end{equation}
where $\mathbb D$ is a similarity measure between policies.
The methods for solving the problem can be divided into three groups, as shown in Fig. \ref{fig:review_of_rl_il} (b), which are reviewed below.

\subsection{Behavior Clone}
Behavior clone (\textbf{BC}) formulates the problem as a supervised learning process with the objective of matching the learned policy $\pi_\theta$ to the expert policy $\pi_E$:
\begin{equation}
\label{BC_objective}
\min_{\theta} \mathbb E ||\pi_\theta - \pi_E||_2
\end{equation}
which is typically realized by minimizing the L2-loss:
\begin{equation}
\label{BC_learning}
J(\theta) = \mathbb E_{(s,a) \sim \cal D}[(\pi_\theta(s)-a)^2]
\end{equation}

Early research on imitation learning can be dated back to the ALVINN system \cite{pomerleau1989alvinn}, which used a 3-layer neural network to perform road following based on front camera images. In the most recent decade, deep imitation learning (DIL) has been conducted using DNNs as policy function approximators and has realized success in end-to-end AD systems \cite{bojarski2016end,bojarski2017explaining,xu2017end}. BC performs well for states that are covered by the training distribution but generalizes poorly to new states due to compounding errors in the actions, which is also referred to as \textbf{covariate shift} \cite{pomerleau1991efficient,ross2010efficient}. To overcome this problem, Ross et al. \cite{ross2011reduction} proposed DAgger, which improves upon supervised learning by using a primary policy to collect training examples while running a reference policy simultaneously. In each iteration, states that are visited by the primary policy are also sent to the reference policy to output expert actions, and the newly generated demonstrations are aggregated into the training dataset. SafeDAgger \cite{zhang2016query} extends on DAgger by introducing a safe policy that learns to predict the error that is made by a primary policy without querying the reference policy. In addition to dataset aggregation, data augmentation techniques such as the addition of random noise to the expert action have also been commonly used in DIL \cite{codevilla2018end,chen2019deep}.
\subsection{Inverse Reinforcement Learning}
The inverse reinforcement learning problem, which was first formulized in the study of Ng et al. \cite{ng2000algorithms}, is to identify a reward function $r_\theta$ for which the expert behavior is optimal:
\begin{equation}
\label{IRL_objective}
\max_\theta \mathbb E_{\pi_E}[G_t|r_\theta] - \mathbb E_\pi [G_t|r_\theta]
\end{equation}

Early studies utilized linear function approximation of reward functions and identified the optimal reward via maximum margin approaches \cite{abbeel2004apprenticeship,ratliff2006maximum}. By introducing the maximum entropy principle, Ziebart et al. \cite{ziebart2008maximum} eliminated the reward ambiguity between demonstrations and expert policy where multiple rewards may explain the expert behavior. The reward function is learned through maximizing the posterior probability of observing expert trajectories:
\begin{equation}
\label{IRL_learning}
J(\theta) = \mathbb E_{\xi_i \sim \cal D}[\log P(\xi_i|r_\theta)]
\end{equation}
where the probability of a trajectory satisfies $P(\xi_i|r_\theta) \propto \exp(r_\theta(\xi_i))$.
Several studies have extended the reward functions to nonlinear formulations through Gaussian processes \cite{levine2011nonlinear} or boosting \cite{ratliff2006boosting,ratliff2009learning}. However, the above methods generally operate on low-dimensional features. The use of rich and expressive function approximators, in the form of neural networks, has been proposed to learn reward functions directly on raw high-dimensional state representations \cite{wulfmeier2015maximum,finn2016guided}. 

A problem that is encountered with IRL is that to evaluate the reward function, a forward reinforcement learning process must be conducted to obtain the corresponding policy, thereby rendering IRL inefficient and computationally expensive. Many early approaches require solving an MDP in the inner loop of each iterative optimization \cite{ng2000algorithms,ng2004autonomous,ratliff2006maximum,ziebart2008maximum}. Perfect knowledge of the system dynamics and an efficient offline solver are needed in these methods, thereby limiting their applications in complex real-world scenarios such as robotic control. Finn et al. \cite{finn2016guided} proposed guided cost learning (GCL), which handles unknown dynamics in high-dimensional complex systems and learns complex neural network cost functions through an efficient sample-based approximation.
\subsection{Generative Adversarial Imitation Learning}
Generative adversarial imitation learning (GAIL) \cite{ho2016generative} directly learns a policy from expert demonstrations while requiring neither the reward design in RL nor the expensive RL process in the inner loop of IRL. GAIL establishes an analogy between imitation learning and generative adversarial networks (GANs) \cite{goodfellow2014generative}. The generator $\pi_\theta$ serves as a policy for imitating expert behavior by matching the state-action distribution of demonstrations, while the discriminator $D_\omega \in (0,1)$ serves as a surrogate reward function for measuring the similarity between generated and demonstration samples. The objective function of GAIL is formulated in the following min-max form:
\begin{equation}
\label {eqn_gail}
\min \limits_{\pi_\theta} \max \limits_{D_\omega} {\mathbb E}_{\pi_\theta} [\log D_\omega(s,a)] + {\mathbb E}_{\pi_E}[\log(1-D_\omega(s,a))] - \lambda H(\pi_\theta)
\end{equation}
where $H(\pi)$ is a regularization entropy term. 
The generator and the discriminator are updated with the following gradients:
\begin{align}
&\nabla_\theta J(\theta) = {\mathbb E}_\pi [\nabla_\theta \log \pi_\theta(a|s)Q(s,a)] - \lambda \nabla_\theta H(\pi_\theta) \nonumber \\
&\nabla_\omega J(\omega) = {\mathbb E}_\pi [\nabla_\omega \log D_\omega(s,a)] + {\mathbb E}_{\pi_E}[\nabla_\omega \log(1-D_\omega(s,a))] 
\end{align}
Finn et al. \cite{finn2016connection} mathematically proved the connection among GANs, IRL and energy-based models. Fu et al.\cite{fu2017learning} proposed adversarial inverse reinforcement learning (AIRL) based on an adversarial reward learning formulation, which can recover reward functions that are robust to dynamics changes.
\section {Architectures of DRL/DIL Integrated AD Systems}
\label{sec:architecture}
\begin{figure*}
\centering
\includegraphics[width=\linewidth]{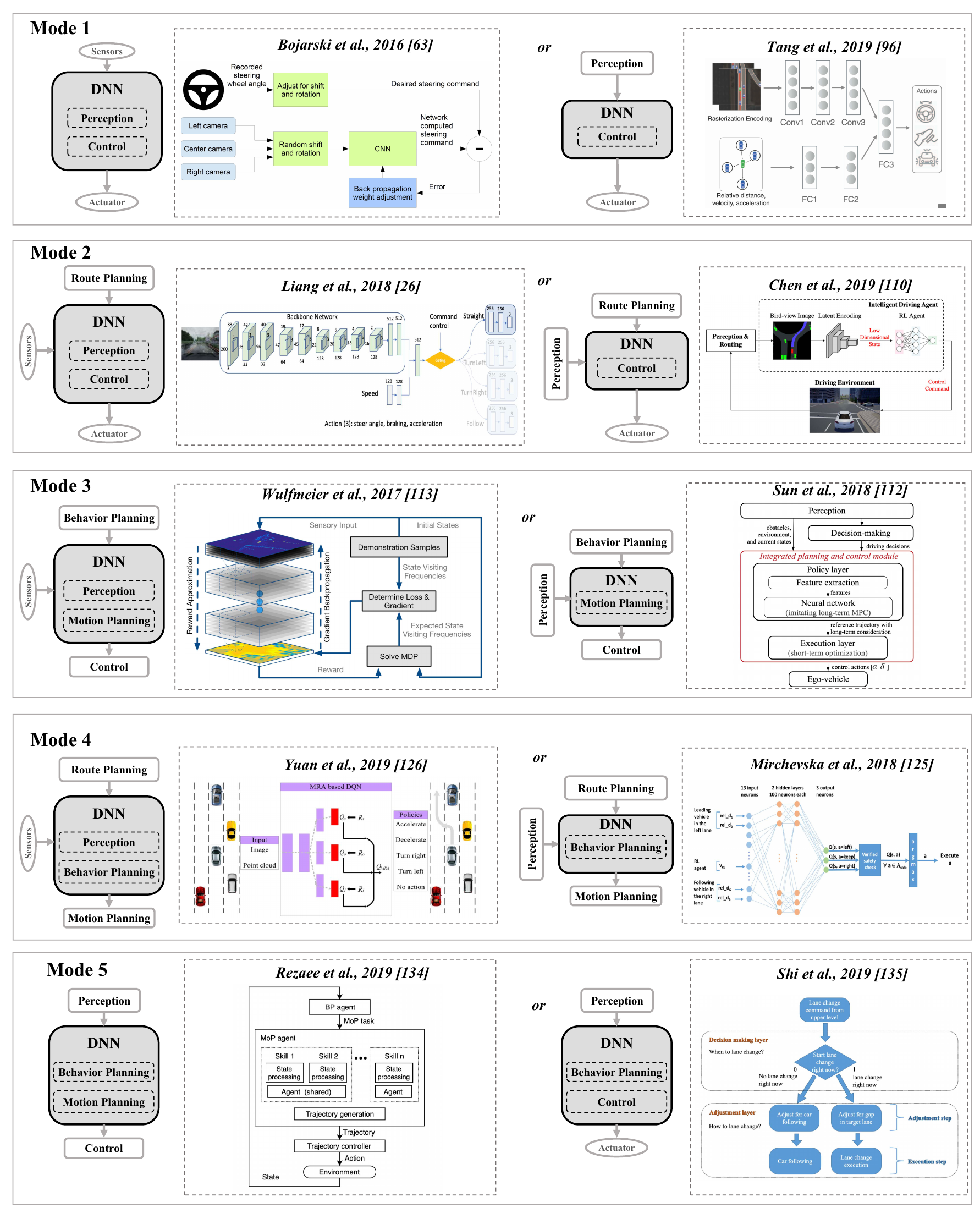}
\caption{Integration modes of
	DRL/DIL models into AD architectures}
\label{fig:modes_all}
\end{figure*}

AD systems have been studied for decades \cite{gonzalez2016review,li2016realtime,paden2016survey,ulbrich2017towards}, which are commonly composed of modular pipelines, as illustrated in Fig. \ref{traditional_architecture}. How can DRL/DIL models be integrated into an AD system and collaborate with other modules? This section reviews the literature from the system architecture perspective, from which five modes are identified, as illustrated in Fig. \ref{fig:modes_all}.
A classification of the studies in each mode is presented in Table \ref{tab:architecture}, along with the exploited DRL/DIL methods, the upstream module for perception, the targeted AD tasks, and the advantages and disadvantages of the architectures, among other information.
Below, we detail each mode of the architectures, which is followed by a comparison of the number of studies that correspond to each mode or to the use of DRL or DIL methods.
\begin{table*}[]
\centering
\large
\caption{A taxonomy of the literature by the integration modes of
	\colorbox{rl}{DRL} or \colorbox{il}{DIL} models into AD architectures}
\label{tab:architecture}
\resizebox{0.9\textwidth}{!}{%
\begin{threeparttable}
\begin{tabular}{lllllcll}
\hline
\multicolumn{1}{c}{\textbf{Architecture}} &
\multicolumn{1}{c}{\textbf{Advantages}} &
\multicolumn{1}{c}{\textbf{Disadvantages}} &
\multicolumn{1}{c}{\textbf{Studies}} &
\multicolumn{1}{c}{\textbf{Methods}} &
\textbf{Perception$^1$} &
\multicolumn{1}{c}{\textbf{Tasks$^2$}} &
\multicolumn{1}{c}{\textbf{Remarks$^3$}} \\ \hline
\multirow{6}{*}{\textbf{\begin{tabular}[c]{@{}l@{}}Mode 1.\\ DRL/DIL\\ Integrated\\ Control\end{tabular}}} &
\multirow{6}{*}{\begin{tabular}[c]{@{}l@{}}- Features a simple\\ structure and adapts \\ to various learning\\ methods.\end{tabular}} &
\multirow{6}{*}{\begin{tabular}[c]{@{}l@{}}- Limited to \\ specified tasks\\ or skills.\\ - Bypassing\\ planning modules\\ weakens the\\ model's\\ interpretability\\ and capability.\end{tabular}} &
\cellcolor{il}
\begin{tabular}[c]{@{}l@{}} \cite{pomerleau1989alvinn}, \cite{muller2006off},\\ \cite{bojarski2016end}, \cite{bojarski2017explaining},\\ \cite{rausch2017learning}, \cite{eraqi2017end},\\ \cite{xu2017end}, \cite{wang2019deep} \\ \end{tabular} &
\cellcolor{il}
BC &
D &
\begin{tabular}[c]{@{}l@{}}road/lane following\\ urban driving\end{tabular} &
\begin{tabular}[c]{@{}l@{}}safety: \cite{zhang2016query}\end{tabular} \\ 
\cmidrule(r){4-8}
&
&
&
\cellcolor{il}
\cite{zhang2016query}, \cite{pan2017agile} &
\cellcolor{il}
DAgger &
D &
road/lane following &
\\ 
\cmidrule(r){4-8}
&
&
&
{\cellcolor{rl}}
\begin{tabular}[c]{@{}l@{}}\cite{bouton2019safe}, \cite{wang2018reinforcement}, \\ \cite{chen2017learning}, \cite{wang2017formulation},\\ \cite{chae2017autonomous}, \cite{bouton2019cooperation}\end{tabular} &
{\cellcolor{rl}}
\begin{tabular}[c]{@{}l@{}}NQL\\ DQN\end{tabular} &
T &
\begin{tabular}[c]{@{}l@{}}lane changing\\ traffic merging\\ imminent events\\ intersection\end{tabular} &
\begin{tabular}[c]{@{}l@{}}safety: \cite{bouton2019safe}\\ interaction:\\ \cite{bouton2019cooperation}\end{tabular} \\ \cmidrule{4-8} 
&
&
&
{\cellcolor{rl}}
\cite{tang2019towards}, \cite{folkers2019controlling} &
{\cellcolor{rl}}
PPO &
T &
\begin{tabular}[c]{@{}l@{}}traffic merging\\ urban driving\end{tabular} &
\\ \cmidrule{4-8} 
&
&
&
\cellcolor{rl}
\cite{kendall2019learning}, \cite{porav2018imminent} &
\cellcolor{rl}
DDPG &
D &
\begin{tabular}[c]{@{}l@{}}road/lane following\\ imminent events\end{tabular} &
\\ \cmidrule{4-8} 
&
&
&
\cellcolor{rl}
\begin{tabular}[c]{@{}l@{}}\cite{wang2018deep}, \cite{wang2019continuous},\\ \cite{kaushik2018overtaking}, \cite{sallab2016end},\\ \cite{vasquez2019multi}\end{tabular} &
\cellcolor{rl}
DDPG &
T &
\begin{tabular}[c]{@{}l@{}}road/lane following\\ lane changing\\ overtaking\\ imminent events\end{tabular} &
\\ \hline \hline
\multirow{6}{*}{\textbf{\begin{tabular}[c]{@{}l@{}}Mode 2.\\ Extension of\\ Mode 1 with\\ High-level\\ Command\end{tabular}}} &
\multirow{6}{*}{\begin{tabular}[c]{@{}l@{}}- Considers both\\ high-level planning\\ and perception\\ - Generates distinct\\ control behaviors\\ according to the \\ high-level decisions\end{tabular}} &
\multirow{6}{*}{\begin{tabular}[c]{@{}l@{}}- Single model\\ may not capture\\ sufficiently diverse\\ behaviors\\ - Learning a model for each\\ behavior \\ increases the training\\ cost and the\\ demand for data \end{tabular}} &
\cellcolor{il}
\cite{hecker2018end} &
\cellcolor{il}
BC &
D &
urban driving &
\\ \cmidrule{4-8} 
&
&
&
\cellcolor{il}
\begin{tabular}[c]{@{}l@{}}\cite{dosovitskiy2017carla}, \cite{codevilla2018end},\\ \cite{abdou2019end}\end{tabular} &
\cellcolor{il}
CIL &
D &
urban navigation &
\begin{tabular}[c]{@{}l@{}}\end{tabular} \\ \cmidrule{4-8} 
&
&
&
\cellcolor{il}
\begin{tabular}[c]{@{}l@{}}\cite{cui2019uncertainty}, \cite{tai2019visual}\end{tabular} &
\cellcolor{il}
UAIL &
D &
urban navigation &
\begin{tabular}[c]{@{}l@{}}uncertainty: \\ \cite{cui2019uncertainty}, \cite{tai2019visual}\end{tabular} \\ \cmidrule{4-8} 
&
&
&
\cellcolor{rl}
\cite{buechel2018deep} &
\cellcolor{rl}
DDPG &
T &
parking &
\\ \cmidrule{4-8} 
&
&
&
\cellcolor{rl}
\cite{liang2018cirl} &
\cellcolor{rl}
DDPG &
D &
urban navigation &
\\ \cmidrule{4-8} 
&
&
&
\cellcolor{rl}
\cite{chen2019model} &
\cellcolor{rl}
\begin{tabular}[c]{@{}l@{}}DDQN/TD3/SAC\end{tabular} &
T &
roundabout &
\\ \hline \hline
\multirow{5}{*}{\textbf{\begin{tabular}[c]{@{}l@{}}Mode 3.\\ DRL/DIL\\ Integrated\\ Motion\\ Planning\end{tabular}}} &
\multirow{5}{*}{\begin{tabular}[c]{@{}l@{}}- Learn to imitate\\ human trajectories\\ - Efficient forward\\ prediction process\end{tabular}} &
\multirow{5}{*}{\begin{tabular}[c]{@{}l@{}}- No guarantee on\\ safety or feasibility\end{tabular}} &
\cellcolor{il}
\begin{tabular}[c]{@{}l@{}}\cite{bansal2018chauffeurnet}, \cite{chen2019deep}\end{tabular} &
\cellcolor{il}
BC &
T &
urban driving &
\begin{tabular}[c]{@{}l@{}} safety: \cite{chen2019deep}\end{tabular} \\ \cmidrule{4-8} 
&
&
&
\cellcolor{il}
\cite{sun2018fast} &
\cellcolor{il}
DAgger &
T &
highway driving &
\\ \cmidrule{4-8} 
&
&
&
\cellcolor{il}
\cite{wulfmeier2017large} &
\cellcolor{il}
MaxEnt DIRL &
D &
urban driving &
\\ \cmidrule{4-8} 
&
&
&
\cellcolor{rl}
\cite{bernhard2018experience} &
\cellcolor{rl}
\begin{tabular}[c]{@{}l@{}}DDQN/DQfD\end{tabular} &
T &
valet parking &
\begin{tabular}[c]{@{}l@{}}safety: \cite{bernhard2018experience} \end{tabular} \\ \cmidrule{4-8} 
&
&
&
\cellcolor{rl}
\cite{hart2019lane} &
\cellcolor{rl}
SAC &
T &
traffic merge &
\\ \hline \hline
\multirow{8}{*}{\textbf{\begin{tabular}[c]{@{}l@{}}Mode 4.\\ DRL/DIL\\ Integrated\\ Behavior\\ Planning\end{tabular}}} &
\multirow{8}{*}{\begin{tabular}[c]{@{}l@{}}- The DNNs need only\\ plan high-\\ level behavioral\\ actions.\end{tabular}} &
\multirow{8}{*}{\begin{tabular}[c]{@{}l@{}}- Simple and few\\ actions limit the \\ control precision\\ and diversity of \\ driving styles.\\ - Complicated and\\ too many actions\\ increase the training\\ cost.\end{tabular}} &
\cellcolor{il}
\cite{wang2019human} &
\cellcolor{il}
AIRL &
T &
lane change &
\\ \cmidrule{4-8} 
&
&
&
\cellcolor{il}
\cite{sharifzadeh2016learning} &
\cellcolor{il}
IRL &
D &
lane change &
\\ \cmidrule{4-8} 
&
&
&
\cellcolor{rl}
\begin{tabular}[c]{@{}l@{}}\cite{alizadeh2019automated}, \cite{deshpande2019deep},\\ \cite{tram2019learning}, \cite{isele2018navigating},\\ \cite{li2019urban}, \cite{ronecker2019deep},\\ \cite{wolf2018adaptive}, \cite{mirchevska2018high}\end{tabular} &
\cellcolor{rl}
DQN &
T &
\begin{tabular}[c]{@{}l@{}}lane change\\ intersection\end{tabular} &
\begin{tabular}[c]{@{}l@{}}\end{tabular} \\ \cmidrule{4-8} 
&
&
&
\cellcolor{rl}
\cite{yuan2019multi}, \cite{lee2019may} &
\cellcolor{rl}
DQN &
D &
lane change &
\begin{tabular}[c]{@{}l@{}}\end{tabular} \\ \cmidrule{4-8} 
&
&
&
\cellcolor{rl}
\cite{you2018highway}, \cite{wang2019q} &
\cellcolor{rl}
Q-Learning &
T &
lane changing &
\\ \cmidrule{4-8} 
&
&
&
\cellcolor{rl}
\cite{hoel2018automated}, \cite{zhang2018human} &
\cellcolor{rl}
DDQN &
T &
\begin{tabular}[c]{@{}l@{}}lane changing\\ urban driving\end{tabular} &
\\ \cmidrule{4-8} 
&
&
&
\cellcolor{rl}
\cite{liu2019learning} &
\cellcolor{rl}
DQfD &
T &
lane changing &
\begin{tabular}[c]{@{}l@{}}safety: \cite{liu2019learning}\end{tabular} \\ \cmidrule{4-8} 
&
&
&
\cellcolor{rl}
\cite{min2019deep} &
\cellcolor{rl}
QR-DQN &
D &
highway driving &
\\ \hline \hline
\multirow{5}{*}{\textbf{\begin{tabular}[c]{@{}l@{}}Mode 5.\\ DRL/DIL\\ Integrated\\ Hierarchical\\ Planning and\\ Control\end{tabular}}} &
\multirow{5}{*}{\begin{tabular}[c]{@{}l@{}}- Simultaneously\\ plan at various\\ levels of abstraction.\end{tabular}} &
\multirow{5}{*}{\begin{tabular}[c]{@{}l@{}}- Hierarchical DNNs\\ increase the training cost\\ and decrease \\ the convergence speed.\end{tabular}} &
\cellcolor{rl}
\cite{rezaee2019multi}, \cite{shi2019driving} &
\cellcolor{rl}
DQN &
T &
\begin{tabular}[c]{@{}l@{}}cruise control\\ lane changing\end{tabular} &
\\ \cmidrule{4-8} 
&
&
&
\cellcolor{rl}
\cite{chen2018deep} &
\cellcolor{rl}
\begin{tabular}[c]{@{}l@{}}Hierarchical\\ Policy gradient\end{tabular} &
T &
traffic light passing &
\\ \cmidrule{4-8} 
&
&
&
\cellcolor{rl}
\cite{chen2019attention} &
\cellcolor{rl}
DDPG &
D &
lane changing &
\begin{tabular}[c]{@{}l@{}}\end{tabular} \\ \cmidrule{4-8} 
&
&
&
\cellcolor{rl}
\cite{qiao2018pomdp} &
\cellcolor{rl}
DDPG &
T &
intersection &
\\ \cmidrule{4-8} 
&
&
&
\cellcolor{rl}
\cite{paxton2017combining} &
\cellcolor{rl}
DDPG &
T &
urban driving &
\begin{tabular}[c]{@{}l@{}}\end{tabular} \\ \hline
\end{tabular}
\begin{tablenotes}
\item[1] Type of upstream perception module: (D)eep learning method/(T)raditional method
\item[2] For detailed information about AD tasks, see Table \ref{general_tasks} 
\item[3] Studies that also address safety \ref{section:safety}, interaction \ref{section:interaction} and uncertainty \ref{section:uncertainty} problems are labelled.
\end{tablenotes}
\end{threeparttable}
}
\vspace{-0.5cm}
\end{table*}
\subsection {Mode 1. DRL/DIL Integrated Control}
Many studies have applied DRL/DIL to control, which can be abstracted as the architecture of Mode 1 and is illustrated in Fig. \ref{fig:modes_all}. 
Bojarski et al. \cite{bojarski2016end} proposed a well-known end-to-end self-driving control framework. They trained a nine-layer CNN by supervised learning to learn the steering policy without explicit manual decomposition of sequential modules. However, their model only adapts to lane keeping. An alternative option is to feed traditional perception results into the DNN control module. Tang et al. \cite{tang2019towards} proposed the use of environmental rasterization encoding, along with the relative distance, velocity and acceleration, as input to a two-branch neural network, which was trained via proximal policy optimization. 

Although Mode 1 features a simple structure and adapts to a large variety of learning methods, it is limited to specified tasks; thus, it has difficulty addressing scenarios in which multiple driving skills that are conditioned on various situations are needed. Moreover, bypassing and ignoring behavior planning or motion planning processes may weaken the model's interpretability and performance in complex tasks (e.g., urban navigation).
\subsection {Mode 2. Extension of Mode 1 with High-level Command}
As illustrated in Fig. \ref{fig:modes_all}, Mode 2 extends Mode 1 by considering the high-level planning output. The control module is composed of either a general model for all behaviors \cite{hecker2018end,chen2019model} or a series of models for distinct behaviors \cite{dosovitskiy2017carla,codevilla2018end,liang2018cirl, abdou2019end, tai2019visual}. Chen et al. \cite{chen2019model} projected detected environment vehicles and the routing onto a bird-view road map, which served as the input of a policy network. Liang et al. \cite{liang2018cirl} built on conditional imitation learning (\textbf{CIL}) \cite{codevilla2018end} and proposed the branched actor network, as illustrated in Fig. \ref{fig:modes_all}. These methods learn several control submodules for distinct behaviors. A gating control command from high-level route planning and behavior planning modules is responsible for the selection of the corresponding control submodule.

Although Mode 2 mitigates the problems that are encountered with Mode 1, it has its own limitations. A general model may be not sufficient for capturing diverse behaviors. However, learning a model for each behavior increases the demand for training data. Moreover, Mode 2 may be not as computationally efficient as Mode 1 since it requires high-level planning modules that are determined in advance to guide the control module.
\subsection {Mode 3. DRL/DIL Integrated Motion Planning}
Mode 3 integrates DRL/DIL into the motion planning module, and its architecture is illustrated in Fig. \ref{fig:modes_all}. Utilizing the planning output (e.g., routes and driving decisions) from high-level modules, along with the current perception output, DNNs are trained to predict future trajectories or paths. DIL models \cite{bansal2018chauffeurnet,sun2018fast,wulfmeier2017large,chen2019deep} are the mainstream choices for implementing this architecture. 
As illustrated in Fig. \ref{fig:modes_all}, Sun et al. \cite{sun2018fast} proposed training a neural network that imitates long-term MPC via Sampled-DAgger, where the policy network's input was from perception (obstacles, environment, and current states) and decision-making (driving decisions). Alternatively, Wulfmeier et al. \cite{wulfmeier2017large} proposed projecting the LiDAR point cloud onto a grid map, which is sent to the DNN. The DNN is responsible for learning a cost function that guides the motion planning. The control part in Mode 3 typically utilizes traditional control techniques such as PID \cite{chen2019deep} or MPC \cite{sun2018fast}. 

One major disadvantage of Mode 3 is that although DNN planned trajectories can imitate human trajectories, their safety and feasibility cannot be guaranteed.

\begin{table*}[]
\centering
\caption{Comparison of the literature by DRL/DIL integration modes}
\label{tab:modes_number_comparison}
\resizebox{0.8\textwidth}{!}{%
\begin{threeparttable}
\begin{tabular}{|c|c|c|c|c|c|c|c|c|}
\hline
\multirow{2}{*}{\textbf{Perception}} &
\multicolumn{2}{c|}{\textbf{\begin{tabular}[c]{@{}c@{}}Control\\ (Modes 1\&2)\end{tabular}}} &
\multicolumn{2}{c|}{\textbf{\begin{tabular}[c]{@{}c@{}}Motion Planning\\ (Mode 3)\end{tabular}}} &
\multicolumn{2}{c|}{\textbf{\begin{tabular}[c]{@{}c@{}}Behavior Planning\\ (Mode 4)\end{tabular}}} &
\multicolumn{2}{c|}{\textbf{\begin{tabular}[c]{@{}c@{}}Hierarchical P. \& C. $^1$\\ (Mode 5)\end{tabular}}} \\ \cline{2-9} 
& DRL & DIL & DRL & DIL & DRL & \multicolumn{1}{l|}{DIL} & DRL & DIL \\ \hline
\textbf{Traditional} & \textbf{15} & 0 & \textbf{2} & \textbf{3} & \textbf{13} & 1 & \textbf{5} & 0 \\ \hline
\textbf{DNN} & 3 & \textbf{16} & 0 & 1 & 3 & 1 & 1 & 0 \\ \hline
\textbf{Subtotal} & \textbf{18 (52.9\%) } & 16 (47.1\%) & 2 (33.3\%) & \textbf{4 (66.7\%)} & \textbf{16 (88.9\%)} & 2 (11.1\%) & \textbf{6 (100\%)} & 0 (0\%) \\ \hline
\textbf{Total} & \multicolumn{2}{c|}{\textbf{34 (53.1\%)}} & \multicolumn{2}{c|}{6 (9.4\%)} & \multicolumn{2}{c|}{18 (28.1\%)} & \multicolumn{2}{c|}{6 (9.4\%)} \\ \hline
\end{tabular}
\begin{tablenotes}
\footnotesize
\item[*] The values in this table are the numbers and percentages of papers in Table \ref{tab:architecture} that belong to the corresponding categories. 
\item[1] Abbreviation for ``Hierarchical Planning and Control'' 
\end{tablenotes}
\end{threeparttable}
}
\end{table*}
\begin{table*}[]
\centering
\caption{A taxonomy of the literature by scenarios and AD tasks}
\label{general_tasks}
\resizebox{0.9\textwidth}{!}{%
\begin{tabular}{|c|c|c|l|l|}
\hline
\multirow{2}{*}{\textbf{Scenario}} &
\multirow{2}{*}{\textbf{AD Task}} &
\multirow{2}{*}{\textbf{Description}} &
\multicolumn{2}{c|}{\textbf{Ref.}} \\ \cline{4-5} 
&
&
&
\multicolumn{1}{c|}{\textbf{DRL Methods}} &
\multicolumn{1}{c|}{\textbf{DIL Methods}} \\ \hline \hline
\multirow{4}{*}{\textbf{Urban}} &
\textbf{Intersection} &
\begin{tabular}[c]{@{}c@{}}Learn to drive through intersections (while interacting \\ and negotiating with other traffic participants).\end{tabular} &
\begin{tabular}[c]{@{}l@{}}DQN \cite{deshpande2019deep,tram2019learning,isele2018navigating,bouton2019safe}\\ cDQN \cite{paxton2017combining}\\ DDPG \cite{qiao2018pomdp,paxton2017combining}\end{tabular} & ---
\\ \cline{2-5} 
&
\textbf{Roundabout} &
\begin{tabular}[c]{@{}c@{}}Learn to drive through roundabouts (while interacting \\ and negotiating with other traffic participants).\end{tabular} &
\begin{tabular}[c]{@{}l@{}}DDQN \cite{chen2019model}\\ TD3 \cite{chen2019model}\\ SAC \cite{chen2019model}\end{tabular} &
Horizon GAIL\cite{behbahani2019learning} \\ \cline{2-5} 
&
\textbf{Urban Navigation} &
\begin{tabular}[c]{@{}c@{}}Learn to drive in urban environments to reach specified \\ objectives by following global routes.\end{tabular} &
DDPG \cite{liang2018cirl} &
\begin{tabular}[c]{@{}l@{}}BC \cite{hecker2018end}\\ CIL \cite{dosovitskiy2017carla,codevilla2018end, abdou2019end}\\ UAIL \cite{cui2019uncertainty,tai2019visual} \end{tabular} \\ \cline{2-5} 
&
\textbf{Urban Driving} &
Learn to drive in urban environments without specified objectives. &
\begin{tabular}[c]{@{}l@{}} DDQN \cite{li2019urban} \\ PPO \cite{chen2019adaptive}\\ Policy gradient \cite{chen2018deep}\end{tabular} &
\begin{tabular}[c]{@{}l@{}}BC \cite{xu2017end,bansal2018chauffeurnet} \\ DIRL \cite{wulfmeier2017large}\end{tabular} \\ \hline \hline
\multirow{5}{*}{\textbf{Highway}} &
\textbf{Lane Change (LC)} &
Learn to decide and perform lane changes. &
\begin{tabular}[c]{@{}l@{}}DQN \cite{alizadeh2019automated,wang2018reinforcement}\\ DDQN \cite{huegle2019dynamic}\\ DDPG \cite{chen2019attention}\end{tabular} &
\begin{tabular}[c]{@{}l@{}}Projection IRL \cite{sharifzadeh2016learning}\\ AIRL \cite{wang2019human}\end{tabular} \\ \cline{2-5} 
&
\textbf{Lane Keep (LK)} &
Learn to drive while maintaining the current lane. &
\begin{tabular}[c]{@{}l@{}}DQN \cite{sallab2016end}\\ DDPG \cite{kendall2019learning}\end{tabular} &
\begin{tabular}[c]{@{}l@{}}BC \cite{rausch2017learning,eraqi2017end}\\ SafeDAgger \cite{zhang2016query}\end{tabular} \\ \cline{2-5} 
&
\textbf{Cruise Control} &
Learn a policy for adaptive cruise control. &
\begin{tabular}[c]{@{}l@{}}NQL \cite{chen2017learning}\\ DQN \cite{rezaee2019multi}\\ Policy gradient \cite{desjardins2011cooperative} \\ Actor-critic \cite{zhao2013supervised,zhao2017model}\end{tabular} & ---
\\ \cline{2-5} 
&
\textbf{Traffic Merging} &
Learn to merge into highways. &
\begin{tabular}[c]{@{}l@{}}DQN \cite{wang2017formulation}\\ PPO \cite{tang2019towards}\end{tabular} & ---
\\ \cline{2-5} 
&
\textbf{Highway Driving} &
\begin{tabular}[c]{@{}c@{}}Learn a general policy for driving on a highway, which \\ may include multiple behaviors such as LC and LK.\end{tabular} &
\begin{tabular}[c]{@{}l@{}}DQN \cite{min2018deep,yuan2019multi}\\ QR-DQN \cite{min2019deep}\end{tabular} &
\begin{tabular}[c]{@{}l@{}}GAIL \cite{kuefler2017imitating}\\ PS-GAIL \cite{bhattacharyya2018multi}\\ MaxEnt IRL \cite{kuderer2015learning} \end{tabular} \\ \hline \hline
\multirow{2}{*}{\textbf{Others}} &
\textbf{Road Following} &
Learn to simply follow one road. & ---
&
\begin{tabular}[c]{@{}l@{}}BC \cite{pomerleau1989alvinn, muller2006off,bojarski2016end,bojarski2017explaining}\\ DAgger \cite{pan2017agile}\end{tabular} \\ \cline{2-5} 
&
\textbf{Imminent Events} &
Learn to avoid or mitigate imminent events such as collisions. &
\begin{tabular}[c]{@{}l@{}}DQN \cite{chae2017autonomous}\\ DDPG \cite{porav2018imminent}\end{tabular} &
RAIL \cite{bhattacharyya2019simulating}\\ \hline
\end{tabular}%
}
\vspace{-0.5cm}
\end{table*}

\vspace{-0.5cm}
\subsection {Mode 4. DRL/DIL Integrated Behavior Planning}
Many studies focus on integrating DRL/DIL into the behavior planning module and deriving high-level driving policies. The corresponding architecture is presented in Fig. \ref{fig:modes_all}, where DNNs decide behavioral actions and the subsequent motion planning and control modules typically utilize traditional methods. Many studies build upon DQN and its variants \cite{alizadeh2019automated,deshpande2019deep,tram2019learning,isele2018navigating,yuan2019multi,li2019urban,wolf2018adaptive,mirchevska2018high,ronecker2019deep}. 
As illustrated in Fig. \ref{fig:modes_all}, Yuan et al. \cite{yuan2019multi} decomposed the action space into five typical behavioral decisions on a highway. 
Compared with DRL, studies that employ DIL to learn high-level policies are limited. A recent study by Wang et al. \cite{wang2019human} proposed the use of augmented adversarial inverse reinforcement learning (AIRL) to learn human-like decision-making on highways, where the action space consists of all possible combinations of lateral and longitudinal decisions. 

In Mode 4, the design of behavioral actions is nontrivial, and one must balance the training cost and the diversity of driving styles. Simple and few behavioral actions limits the control precision and diversity of driving styles, whereas many and sophisticated actions increases the training cost.

\subsection {Mode 5. DRL/DIL Integrated Hierarchical Planning and Control}
\begin{table*}[]
\centering
\caption{Inputs of the DRL/DIL methods for AD tasks}
\label{tab:input}
\resizebox{0.9\textwidth}{!}{%
\begin{tabular}{|c|c|c|c|c|}
\hline
\multirow{2}{*}{\textbf{\begin{tabular}[c]{@{}c@{}}Information Source\end{tabular}}} &
\multirow{2}{*}{\textbf{Class}} &
\multirow{2}{*}{\textbf{Inputs}} &
\multicolumn{2}{c|}{\textbf{Ref.}} \\ \cline{4-5} 
&
&
&
\textbf{DRL Methods} &
\textbf{DIL Methods} \\ \hline
\multirow{3}{*}{\textbf{Ego Vehicle}} &
\textbf{\begin{tabular}[c]{@{}c@{}}Position Information\end{tabular}} &
ego position &
\cite{chen2019model,tram2019learning,bouton2019safe,chen2018deep,li2019urban,wang2017formulation,wang2018reinforcement} &
\cite{bhattacharyya2018multi,kuefler2017imitating,bhattacharyya2019simulating,bansal2018chauffeurnet} \\ \cline{2-5} 
&
\textbf{\begin{tabular}[c]{@{}c@{}}Heading Information\end{tabular}} &
\begin{tabular}[c]{@{}c@{}}heading angle, orientation,\\ steering, yaw, and yaw rate\end{tabular} &
\cite{deshpande2019deep,isele2018navigating,wang2017formulation,wang2018reinforcement,kendall2019learning,paxton2017combining} &
\cite{bhattacharyya2018multi,xu2017end,bojarski2017explaining,kuefler2017imitating,bhattacharyya2019simulating,behbahani2019learning} \\ \cline{2-5} 
&
\textbf{\begin{tabular}[c]{@{}c@{}}Speed Information\end{tabular}} &
speed/velocity and acceleration &
\begin{tabular}[c]{@{}c@{}}\cite{paxton2017combining, alizadeh2019automated, li2019urban, wang2018reinforcement,huegle2019dynamic,deshpande2019deep,tram2019learning,isele2018navigating,qiao2018pomdp,bouton2019safe,chen2018deep,liang2018cirl}\\ \cite{wang2017formulation,sallab2016end,kendall2019learning,huang2017parameterized,chae2017autonomous,zhao2017model,rezaee2019multi,porav2018imminent}\end{tabular} &
\begin{tabular}[c]{@{}c@{}}\cite{behbahani2019learning,sharifzadeh2016learning,cui2019uncertainty,tai2019visual,abdou2019end,dosovitskiy2017carla,codevilla2018end,xu2017end}\\ \cite{bhattacharyya2018multi,kuderer2015learning,kuefler2017imitating,bhattacharyya2019simulating,pan2017agile}\end{tabular} \\ \hline
\multirow{7}{*}{\textbf{Road Environment}} &
\multirow{3}{*}{\textbf{Pixel Data}} &
camera RGB images &
\cite{yuan2019multi,min2018deep,chen2019attention,min2019deep,liang2018cirl,kendall2019learning} &
\begin{tabular}[c]{@{}c@{}}\cite{zhang2016query,rausch2017learning,eraqi2017end,cui2019uncertainty,tai2019visual,abdou2019end,dosovitskiy2017carla,codevilla2018end,xu2017end,hecker2018end}\\ \cite{pomerleau1989alvinn,yu1995road,muller2006off,pan2017agile,bojarski2016end,bojarski2017explaining}\end{tabular} \\ \cline{3-5} 
&
&
\begin{tabular}[c]{@{}c@{}}semantically segmented images\end{tabular} &
\cite{porav2018imminent} &
--- \\ \cline{3-5} 
&
&
2D bird’s-eye-view images &
\cite{tang2019towards} &
--- \\ \cline{2-5} 
&
\multirow{2}{*}{\textbf{Point Data}} &
LiDAR sensor readings &
\cite{yuan2019multi,min2018deep,min2019deep} &
\cite{bhattacharyya2018multi,kuefler2017imitating,bhattacharyya2019simulating,behbahani2019learning} \\ \cline{3-5} 
&
&
2D LiDAR grid map &
--- &
\cite{wulfmeier2017large} \\ \cline{2-5} 
&
\multirow{2}{*}{\textbf{Object Data}} &
\begin{tabular}[c]{@{}c@{}}other road users’ information:\\ relative speed, position, \\ and distance to ego\end{tabular} &
\begin{tabular}[c]{@{}c@{}}\cite{huegle2019dynamic,alizadeh2019automated,deshpande2019deep,chen2019model,tram2019learning,isele2018navigating,li2019urban,paxton2017combining}\\ \cite{wang2017formulation,rezaee2019multi,desjardins2011cooperative,chae2017autonomous,chen2017learning,zhao2017model,zhao2013supervised,tang2019towards}\end{tabular} &
\cite{bhattacharyya2018multi,bhattacharyya2019simulating,wang2019human,bansal2018chauffeurnet} \\ \cline{3-5} 
&
&
\begin{tabular}[c]{@{}c@{}}lane/road information:\\ ego vehicle's distance to\\ lane markings, road\\ curvature, and lane width\end{tabular} &
\cite{alizadeh2019automated,rezaee2019multi,qiao2018pomdp,li2019urban,wang2017formulation,wang2018reinforcement,sallab2016end,huegle2019dynamic} &
\cite{bhattacharyya2018multi,kuefler2017imitating,bhattacharyya2019simulating,wang2019human} \\ \hline
\multirow{3}{*}{\textbf{Task}} &
\textbf{\begin{tabular}[c]{@{}c@{}}Navigation Information\end{tabular}} &
\begin{tabular}[c]{@{}c@{}}navigational driving commands \\ or planned routes\end{tabular} &
\cite{chen2019model,liang2018cirl} &
\cite{abdou2019end,dosovitskiy2017carla,codevilla2018end,hecker2018end,bansal2018chauffeurnet} \\ \cline{2-5} 
&
\textbf{\begin{tabular}[c]{@{}c@{}}Destination Information\end{tabular}} &
\begin{tabular}[c]{@{}c@{}}destination position, distance \\ or angle to destination\end{tabular} &
--- &
\cite{behbahani2019learning} \\ \cline{2-5} 
&
\textbf{\begin{tabular}[c]{@{}c@{}}Traffic Rule Information\end{tabular}} &
\begin{tabular}[c]{@{}c@{}}traffic lights' state, speed \\ limit, and desired speed\end{tabular} &
\cite{chen2018deep} &
\cite{kuderer2015learning,bansal2018chauffeurnet} \\ \hline 
\textbf{\begin{tabular}[c]{@{}c@{}}Prior Knowledge\end{tabular}} &
\textbf{Road Map} &
\begin{tabular}[c]{@{}c@{}}2D top-down road map images\end{tabular} &
\cite{chen2019model} &
\cite{bansal2018chauffeurnet} \\ \hline
\end{tabular}%
}
\vspace{-0.5cm}
\end{table*}
\begin{figure*}[]
\centering
\subfigure[]{
\begin{minipage}[b]{0.45\textwidth}
\centering
\includegraphics[width=\linewidth]{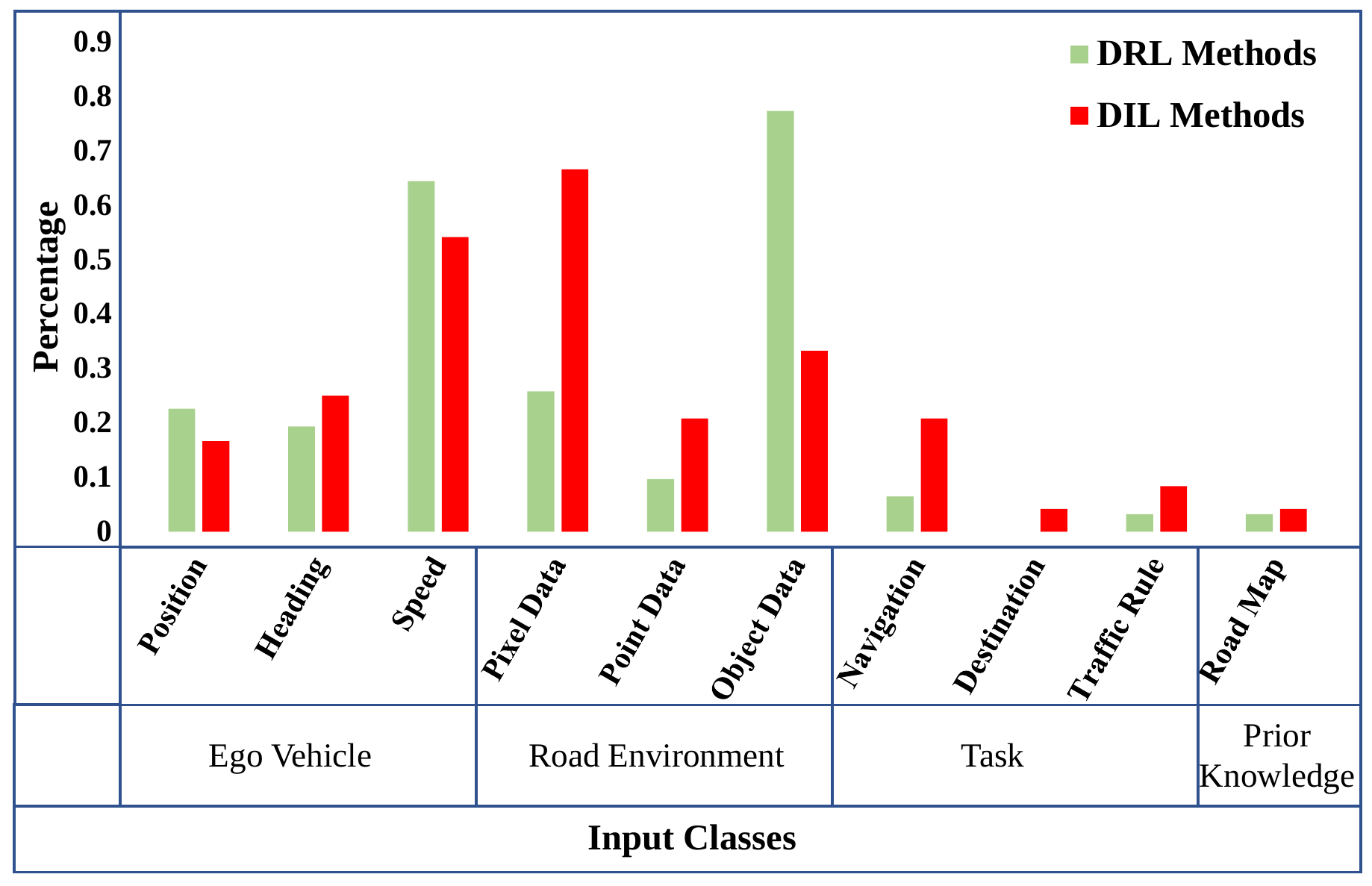}
\vspace{-0.5cm}
\label{fig:input_methods} 
\end{minipage}
} 
\subfigure[]{
\begin{minipage}[b]{0.45\textwidth}
\centering
\includegraphics[width=\linewidth]{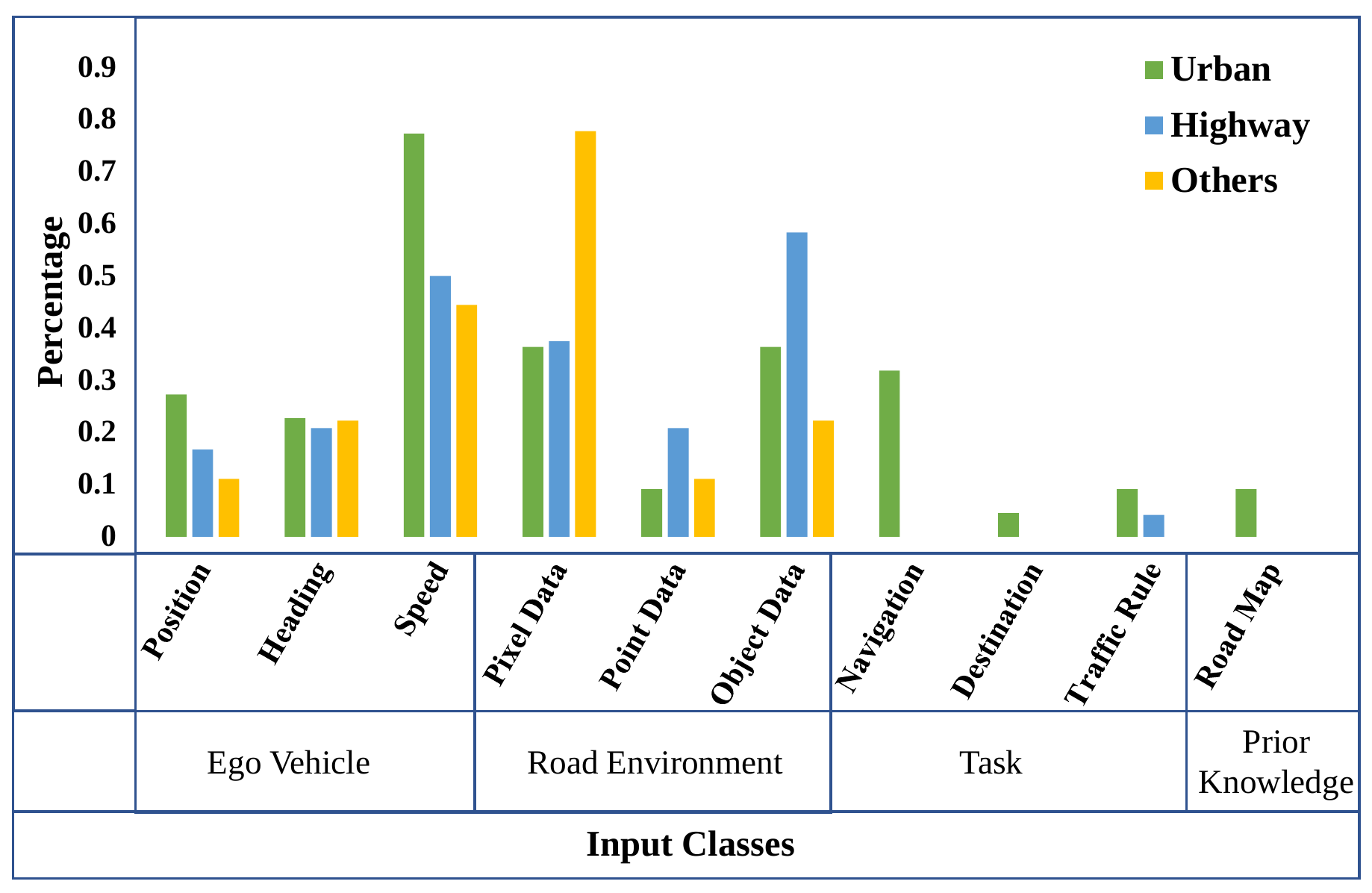}
\vspace{-0.5cm}
\label{fig:input_scenarios}
\end{minipage}
}%
\caption{The percentages of the literature that uses certain data as input. (a) Comparison by DRL/DIL methods. (b) Comparison by scenarios.}
\vspace{-0cm}
\end{figure*}
As illustrated in Fig. \ref{fig:modes_all}, Mode 5 blurs the lines between planning and control, where a single DNN outputs hierarchical actions \cite{chen2019attention} based on the parameterized action space \cite{hausknecht2016deep} or hierarchical DNNs output actions at multiple levels \cite{chen2018deep,rezaee2019multi,shi2019driving}. Rezaee et al. \cite{rezaee2019multi} proposed an architecture, which is illustrated in Fig. \ref{fig:modes_all}, in which BP (behavior planning) is used to make high-level decisions regarding transitions between discrete states and MoP (motion planning) generates a target trajectory according to the decisions that are made via BP. Qiao et al. \cite{qiao2018pomdp} built upon hierarchical MDP (HMDP) and realized their model through an options framework. In their implementation, the hierarchical options were modeled as high-level decisions (SlowForward or Forward). Based on high-level decisions and current observations, low-level control policies were derived.

Mode 5 simultaneously plans at multiple levels, and the low-level planning process considers high-level decisions. However, the use of hierarchical DNNs results in the increase of the training cost and potentially the decrease of the convergence speed since one poorly trained high-level DNN may mislead and disturb the learning process of low-level DNNs. 

\subsection{Statistical Comparison}
A statistical comparison among modes in terms of the number of studies is presented in Table \ref{tab:modes_number_comparison}. Current studies on architectures are premature, unsystematic and unbalanced: 
\begin{itemize}
\item Most studies focus on integrating DRL/DIL into control (Mode1\&2), followed by behavior planning (Mode 4).
\item For Mode 1\&2, DRL methods mainly adopt traditional method-based perception, while DNN-based perception is preferred in DIL methods.
\item DRL seems to be more popular for high-level decision making (Mode 4\&5), while DIL is chosen more frequently for low-level control (Mode 1\&2).
\end{itemize}
Future studies may address the imbalance problem and identify potential new architectures.

\section {Task-Driven Methods}
\label{sec: task-driven}
DRL/DIL studies in AD can be categorized according to their application scenarios and targeted tasks, as listed in Table \ref{general_tasks}. These task-driven studies use DRL/DIL to solve specified AD tasks, where the formulations of DRL/DIL can be decomposed into several key components: 1) \textit{state space and input design}, 2) \textit{action space and output design}, and 3) \textit{reinforcement learning reward design}.
\begin{table*}[]
\centering
\caption{Actions of the DRL/DIL methods for AD tasks}
\label{tab:output}
\resizebox{0.9\textwidth}{!}{%
\begin{tabular}{|c|c|c|c|c|}
\hline
\multirow{2}{*}{\textbf{Action Category}} &
\multirow{2}{*}{\textbf{Subclass}} &
\multirow{2}{*}{\textbf{Action Outputs}} &
\multicolumn{2}{c|}{\textbf{Ref}} \\ \cline{4-5} 
&
&
&
\textbf{DRL Methods} &
\textbf{DIL Methods} \\ \hline
\multirow{4}{*}{\textbf{\begin{tabular}[c]{@{}c@{}}Behavior-level\\ Actions\end{tabular}}} &
\textbf{\begin{tabular}[c]{@{}c@{}}Acceleration \\ Related\end{tabular}} &
e.g., full brake, decelerate, continue, and accelerate &
\cite{deshpande2019deep,isele2018navigating,li2019urban,desjardins2011cooperative,min2018deep,yuan2019multi,min2019deep} & ---
\\ \cline{2-5} 
&
\textbf{\begin{tabular}[c]{@{}c@{}}Lane Change\\ Related\end{tabular}} &
\begin{tabular}[c]{@{}c@{}}e.g., keep, LLC, and RLC; \\ choice of lane change gaps\end{tabular} &
\cite{li2019urban,huegle2019dynamic,alizadeh2019automated, min2018deep,min2019deep} &
\cite{wang2019human} \\ \cline{2-5} 
&
\textbf{\begin{tabular}[c]{@{}c@{}}Turn\\ Related\end{tabular}} &
e.g., straight, left-turn, right-turn, and stop &
\cite{yuan2019multi} &
\cite{sharifzadeh2016learning,xu2017end} \\ \cline{2-5} 
&
\textbf{\begin{tabular}[c]{@{}c@{}}Interaction\\ Related\end{tabular}} &
e.g., take/give way and follow a vehicle; wait/go &
\cite{tram2019learning,isele2018navigating} & ---
\\ \hline
\textbf{\begin{tabular}[c]{@{}c@{}}Trajectory-level\\ Actions\end{tabular}} &
\textbf{\begin{tabular}[c]{@{}c@{}}Planned\\ Trajectory\end{tabular}} &
future path 2D points & ---
&
\cite{wulfmeier2017large,bansal2018chauffeurnet} \\ \hline
\multirow{9}{*}{\textbf{\begin{tabular}[c]{@{}c@{}}Control-level\\ Actions\end{tabular}}} &
\multirow{3}{*}{\textbf{Lateral}} &
discrete steer angles &
\cite{yu1995road} &
\cite{pomerleau1989alvinn} \\ \cline{3-5} 
&
&
continuous steer & ---
&
\cite{eraqi2017end,rausch2017learning,muller2006off,bojarski2016end,bojarski2017explaining} \\ \cline{3-5} 
&
&
\begin{tabular}[c]{@{}c@{}}continuous angular speed\\ or yaw acceleration\end{tabular} &
\cite{wang2018reinforcement} &
\cite{xu2017end} \\ \cline{2-5} 
&
\multirow{3}{*}{\textbf{Longitudinal}} &
discrete acceleration values &
\cite{bouton2019safe, chae2017autonomous} & ---
\\ \cline{3-5} 
&
&
continuous acceleration &
\cite{chen2017learning, zhao2013supervised, zhao2017model} & ---
\\ \cline{3-5} 
&
&
continuous brake and throttle &
\cite{huang2017parameterized} & ---
\\ \cline{2-5} 
&
\multirow{3}{*}{\textbf{\begin{tabular}[c]{@{}c@{}}Simultaneous\\ Lateral \& Longitudinal\end{tabular}}} &
\begin{tabular}[c]{@{}c@{}}continuous steer/turn-rate and\\ speed/acceleration/throttle\end{tabular} &
\cite{wang2017formulation,chen2019model,tang2019towards,kendall2019learning} &
\cite{codevilla2018end,hecker2018end,abdou2019end,pan2017agile,cui2019uncertainty,kuefler2017imitating,bhattacharyya2018multi} \\ \cline{3-5} 
&
&
\begin{tabular}[c]{@{}c@{}}continuous steer, \\ acceleration/throttle and brake\end{tabular} &
\cite{sallab2016end,liang2018cirl,porav2018imminent} &
\cite{dosovitskiy2017carla,tai2019visual} \\ \cline{3-5} 
&
&
continuous steer and binary brake decision & ---
&
\cite{zhang2016query} \\ \hline
\multirow{2}{*}{\textbf{Hierarchical Actions}} &
\textbf{Behavior \& Control} &
\begin{tabular}[c]{@{}c@{}}e.g., behavior-level pass/stop, \\ control-level acceleration, and steer\end{tabular} &
\cite{paxton2017combining,qiao2018pomdp,chen2018deep,chen2019attention} & ---
\\ \cline{2-5} 
&
\textbf{Behavior \& Trajectory} &
\begin{tabular}[c]{@{}c@{}}e.g., behavior-level maintenance, LLC, RLC\\ and trajectory-level path points\end{tabular} &
\cite{rezaee2019multi} & ---
\\ \hline
\end{tabular}%
}
\end{table*}
\begin{table*}
\centering
\caption{Rewards of the DRL methods for AD tasks}
\label{tab:task_reward}
\resizebox{0.9\textwidth}{!}{%
\begin{tabular}{|c|c|c|c|}
\hline
\textbf{Category} &
\textbf{Subclass} &
\textbf{Description} &
\textbf{Ref.} \\ \hline
\multirow{5}{*}{\textbf{Safety}} &
\textbf{Avoid Collision} &
Impose penalties if a collision occurs &
\begin{tabular}[c]{@{}c@{}}\cite{alizadeh2019automated, liang2018cirl,dosovitskiy2017carla,deshpande2019deep, li2019urban, wang2019human,chen2019model,tram2019learning,isele2018navigating,qiao2018pomdp,bouton2019safe,paxton2017combining} \\ \cite{chae2017autonomous,zhao2013supervised,bhattacharyya2019simulating,tang2019towards,zhang2016query,yuan2019multi,min2018deep,min2019deep,porav2018imminent} \end{tabular} \\ \cline{2-4} 
&
\textbf{Time to Collision} &
\begin{tabular}[c]{@{}c@{}}Impose penalties if the time to collision (TTC) is below \\ a safe threshold\end{tabular} &
\cite{alizadeh2019automated,deshpande2019deep,tram2019learning,li2019urban,zhao2017model} \\ \cline{2-4} 
&
\textbf{Distance to Other Vehicles} &
\begin{tabular}[c]{@{}c@{}}Impose penalties if this distance is shorter than a safe threshold\end{tabular} &
\cite{rezaee2019multi,desjardins2011cooperative,zhao2017model,wang2017formulation} \\ \cline{2-4} 
&
\textbf{Number of Lane Changes} &
\begin{tabular}[c]{@{}c@{}}Impose penalties if the number of lane changes is too \\ large or reward a smaller number of lane changes\end{tabular} &
\cite{alizadeh2019automated,rezaee2019multi,yuan2019multi,min2018deep,min2019deep,huegle2019dynamic} \\ \cline{2-4} 
&
\textbf{Out of Road} &
Impose penalties on driving out of road &
\cite{alizadeh2019automated,chen2019attention,sallab2016end,yu1995road,tang2019towards} \\ \hline
\multirow{5}{*}{\textbf{Efficiency}} &
\textbf{Speed} &
\begin{tabular}[c]{@{}c@{}}Reward higher speed until the maximum speed limit is reached;\\ Impose penalties if the speed is lower than the minimum speed limit\end{tabular} &
\begin{tabular}[c]{@{}c@{}}\cite{paxton2017combining,alizadeh2019automated,huegle2019dynamic,deshpande2019deep,li2019urban,chen2019model,chen2018deep,liang2018cirl,dosovitskiy2017carla,chen2019adaptive}\\ \cite{min2019deep,yuan2019multi,min2018deep,rezaee2019multi,wang2017formulation,sallab2016end,kendall2019learning,tang2019towards,chen2019attention,huang2017parameterized}\end{tabular} \\ \cline{2-4} 
&
\textbf{Success} &
Reward the agent if it finished the task successfully &
\cite{alizadeh2019automated,tram2019learning,isele2018navigating,qiao2018pomdp,bouton2019safe,zhao2013supervised,wang2019human,tang2019towards,paxton2017combining} \\ \cline{2-4} 
&
\textbf{Number of Overtakes} &
Reward a higher number of overtakes for efficiency &
\cite{yuan2019multi,min2018deep,chen2019attention,min2019deep} \\ \cline{2-4} 
&
\textbf{Time} &
\begin{tabular}[c]{@{}c@{}}Impose a negative reward in each step to encourage the agent finish \\ the task faster or penalize the agent if the task cannot \\ be finished within a time threshold\end{tabular} &
\cite{chen2019model,isele2018navigating,wang2018reinforcement,qiao2018pomdp} \\ \cline{2-4} 
&
\textbf{Distance to the Destination} &
Provide a larger reward the closer the agent is to the destination &
\cite{qiao2018pomdp,chen2018deep,dosovitskiy2017carla} \\ \hline
\textbf{Comfort} &
\textbf{Jerk} &
\begin{tabular}[c]{@{}c@{}}Impose penalties if the longitudinal or lateral control \\ is too urgent\end{tabular} &
\cite{chen2019model,tram2019learning,chen2018deep,wang2017formulation,wang2018reinforcement,huang2017parameterized,zhao2017model,bhattacharyya2019simulating,tang2019towards,paxton2017combining} \\ \hline
\multirow{4}{*}{\textbf{\begin{tabular}[c]{@{}c@{}}Traffic\\ Rules\end{tabular}}} &
\textbf{Lane Mark Invasion} &
Impose penalties if the agent invades the lane marks &
\cite{liang2018cirl,dosovitskiy2017carla,bhattacharyya2019simulating} \\ \cline{2-4} 
&
\textbf{Distance to the Lane Centerlines} &
\begin{tabular}[c]{@{}c@{}}Impose penalties if the agent deviates from the lane\\ centerlines or routing baselines\end{tabular} &
\cite{chen2019model,chen2019adaptive,chen2019attention,yu1995road,tang2019towards,paxton2017combining} \\ \cline{2-4} 
&
\textbf{Wrong Lane} &
\begin{tabular}[c]{@{}c@{}}Impose penalties if the agent is in the wrong lane, e.g., \\ staying in the left-turn lane if the assigned route is straight\end{tabular} &
\cite{li2019urban} \\ \cline{2-4} 
&
\textbf{Blocking Traffic} &
\begin{tabular}[c]{@{}c@{}}Impose penalties if the agent blocks the future paths of other\\ vehicles that have the right of way\end{tabular} &
\cite{qiao2018pomdp} \\ \hline
\end{tabular}%
}
\end{table*}
\begin{figure*}[]
	\centering
	\includegraphics[width=0.85\linewidth]{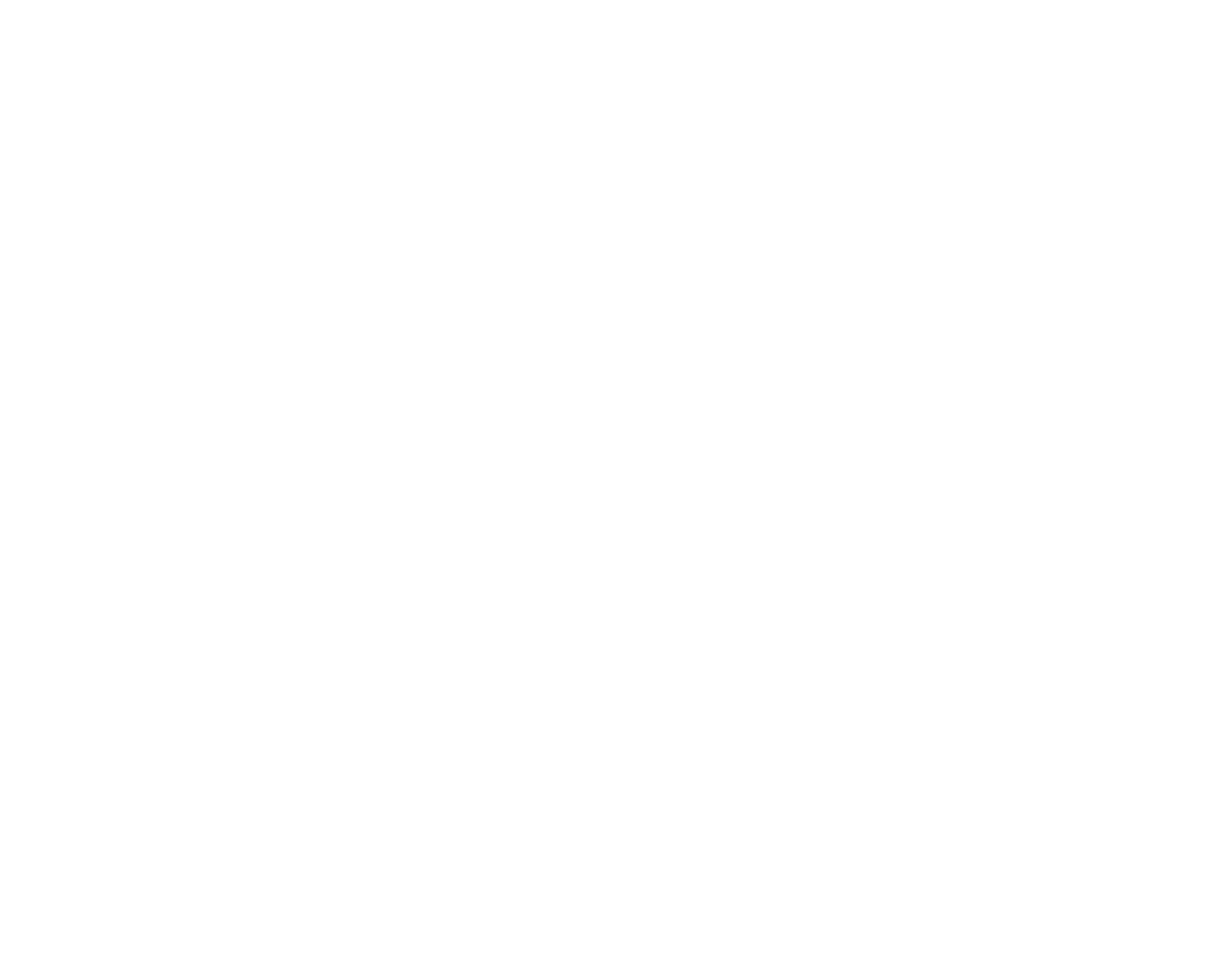}
	\caption{A taxonomy of the literature on how driving safety is addressed by DRL/DIL models. (a)-(c) Three main methods with typical examples in literature.}
	\label{safety_tree_and_examples} 
	\vspace{-0.6cm}
\end{figure*} 
\begin{figure}[b]
	\centering
	\includegraphics[width=0.75\linewidth]{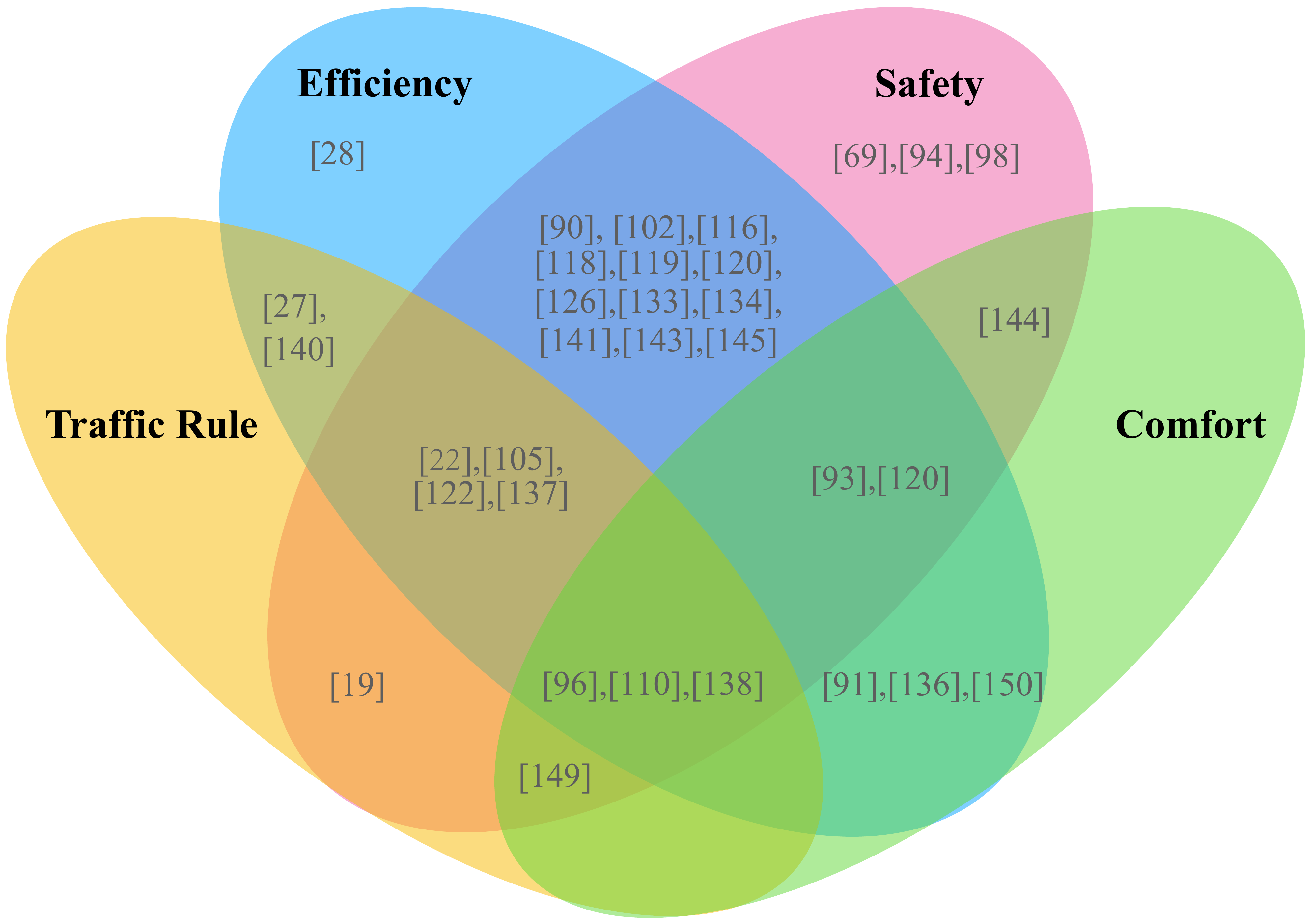}
	\caption{Reinforcement learning rewards for AD tasks.}
	\label{task_reward}
	\vspace{-0.6cm}
\end{figure}
\vspace{-0.5cm}
\subsection {State Space and Input Design}
Table \ref{tab:input} classifies commonly used inputs according to information source (ego vehicle/road environment/task/prior knowledge). A statistical comparison of the percentages of input classes that are used in DRL/DIL methods is presented in Fig. \ref{fig:input_methods}. Compared with the task and prior knowledge, the ego vehicle and road environment seem to be more popular information sources. In all input classes, the ego vehicle speed, pixel data (e.g., camera images) and object data (e.g., other road users' relative speeds and positions) are the most commonly used. Another significant difference between DRL and DIL is that DRL models prefer object data while DIL models prefer pixel data. The selection of low-dimensional object data rather than high-dimensional pixel data as input for DRL models renders the problem more tractable and accelerates the training procedure. Fig. \ref{fig:input_scenarios} presents a statistical comparison of preferences for input classes in various scenarios. Input from the task (e.g., goal positions) and prior knowledge (e.g., road maps) are mostly used in urban scenarios. Point data (e.g., LiDAR sensor readings) and object data are more commonly used in highway scenarios.

Aside from deciding the choice of input, the application of a dynamic input size is an important problem since the number of cars or pedestrians in the ego's vicinity varies over time. Unfortunately, standard DRL/DIL methods rely on inputs of fixed size. The use of occupancy grid maps as inputs of CNNs is a practical solution \cite{isele2018navigating,deshpande2019deep}. However, this solution imposes a trade-off between computational workload and expressiveness. Low-resolution grids decrease the computational burden at the cost of being imprecise in their representation of the environment, whereas for high-resolution grids, most computations may be redundant due to sparsity of the grid maps. Furthermore, a grid map is still limited by its defined size, and agents outside this region is neglected. Alternatively, Everett et al. \cite{everett2018motion} proposed to leverage LSTMs' ability to encode a variable number of agents' observations. Huegle et al. \cite{huegle2019dynamic} suggested the use of deep sets \cite{zaheer17deepsets} as a flexible and permutation-invariant architecture to handle dynamic input. Dynamic input remains an open topic for future studies.
\vspace{-0.4cm}

\subsection {Action Space and Output Design}

A self-driving DRL/DIL agent can plan at different levels of abstraction, namely, low-level control, high-level behavioral planning and trajectory planning, or even at multiple levels simultaneously. According to this, Table \ref{tab:output} categorizes mainstream action spaces into four groups: behavior-level actions, trajectory-level actions, control-level actions and hierarchical actions.
Behavior-level actions are usually designed according to specified tasks. For speed control, acceleration-related actions (e.g., full brake, decelerate, continue, and accelerate \cite{deshpande2019deep}) are commonly used. Lane change actions (e.g., keep/LLC/LRC) and turn actions (e.g., turn left/right/go straight) are preferred in highway scenarios and urban scenarios, respectively. Trajectory-level actions refer to the planned or predicted trajectories/paths \cite{wulfmeier2017large, bansal2018chauffeurnet}, which are typically composed of future path 2D points. Control-level actions refer to low-level control commands (e.g., steer, acceleration, throttle, and brake), which are divided into three classes: lateral control, longitudinal control and simultaneous lateral and longitudinal control. Early studies focused mainly on discrete lateral \cite{pomerleau1989alvinn,yu1995road} or discrete longitudinal control \cite{zhao2013supervised}, while continuous control was considered later \cite{wang2018reinforcement,muller2006off,bojarski2016end,bojarski2017explaining}. Continuous control is demonstrated in \cite{sallab2016end} to produce smoother trajectories than discrete control for lane keeping, which may make passengers feel more comfortable. Simultaneous lateral and longitudinal control has received wide attention, especially in urban scenarios \cite{dosovitskiy2017carla,codevilla2018end,abdou2019end,cui2019uncertainty,tai2019visual}. Recently, hierarchical actions have attracted more attention \cite{paxton2017combining,qiao2018pomdp,chen2018deep,chen2019attention,rezaee2019multi}, which provide higher robustness and interpretability.

\subsection {Reinforcement Learning Reward Design}

A major problem that limits RL's real-world AD applications is the lack of underlying reward functions. Further, the ground truth reward, if it exists, may be multi-modal since human drivers change objectives according to the circumstances. To simplify the problem, current DRL models for AD tasks commonly formulate the reward function as a linear combination of factors, as presented in Fig. \ref{task_reward}. A large proportion of studies consider safety and efficiency. Reward terms that are used in the literature are listed in Table \ref{tab:task_reward}. 
Collison and speed are the most common reward terms when considering safety and efficiency, respectively. 
However, empirically designed reward functions rely heavily on expert knowledge. It is difficult to balance rewards terms, which affects the trained policy performance. Recent studies on predictive reward and multi-reward RL may inspire future investigation. Hayashi et al. \cite{hayashi2019predictive} proposed a predictive reward function that is based on the prediction error of a deep predictive network that models the transition of the surrounding environment. Their hypothesis is that the movement of surrounding vehicles becomes unpredictable when the ego vehicle performs an unnatural driving behavior. Yuan et al. \cite{yuan2019multi} decomposed a single reward function into multi-reward functions to better represent multi-dimensional driving policies through a branched version of Q networks.

\section{Problem-driven Methods}
\label{sec:problem-driven}
AD application has special requirements on factors such as driving safety, interaction with other traffic participants and uncertainty of the environment. This section reviews the literature from the problem-driven perspective with the objectives of determining how these critical issues are addressed by the DRL/DIL models and identifying the challenges that remain.

\vspace{-0.3cm}
\subsection {Safety-enhanced DRL/DIL for AD}
\label{section:safety}
\begin{figure*}[h]
	\centering
	\includegraphics[width=0.85\linewidth]{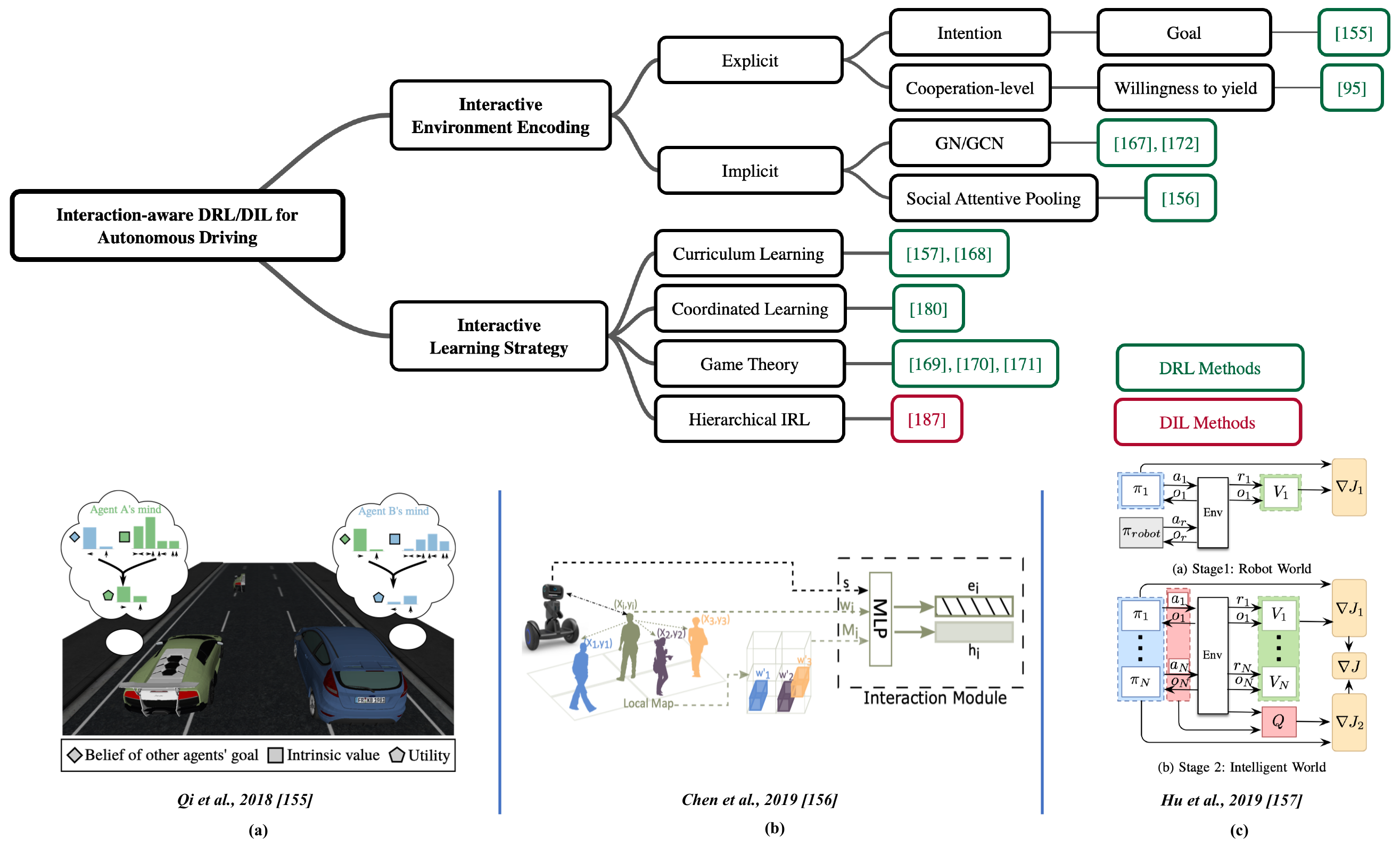}
	\caption{A taxonomy of the literature on interaction-aware DRL/DIL models for AD. (a) Qi et al. \cite{qi2018intent}, (b) Chen et al. \cite{chen2019crowd} are examples of explicit and implicit interactive environment encoding, respectively. (c) Hu et al. \cite{hu2019interaction} is an example of interactive learning strategy.}
	\label{interaction_tree_and_examples} 
	\vspace{-0.6cm}
\end{figure*}

Although DRL/DIL can learn driving policies for complex high-dimensional problems, they only guarantee optimality of the learned policies in a statistical sense. However, in safety-critical AD systems, one failure (e.g., collision) would cause catastrophe. Below, we review representative methods for enhancing safety of DRL/DIL in the AD literature. 
Fig. \ref{safety_tree_and_examples} categorizes the methods into three groups: (a) \textit{modified methods}: methods that modify the original DRL/DIL algorithms, (b) \textit{combined methods}: methods that combine DRL/DIL with traditional methods, and (c) \textit{hybrid methods}: methods that integrate DRL/DIL into traditional methods. 

\subsubsection {Modified Methods}
As illustrated in Fig. \ref{safety_tree_and_examples}(a), modified methods modify the standard DRL/DIL algorithms to enhance safety, typically by constraining the exploration space \cite{garcia15comprehensive,jansen2018shielded,fulton2018safe}.
A safety model checker is introduced to identify the set of actions that satisfy the safety constraints at each state. This can be realized through several approaches, such as goal reachability \cite{bouton2019safe}\cite{bouton2019reinforcement}, probabilistic prediction \cite{isele2018safe} and prior knowledge \& constraints \cite{mukadam2017tactical}. Bouton et al. \cite{bouton2019safe}\cite{bouton2019reinforcement} use a probabilistic model checker, as illustrated in Fig. \ref{safety_tree_and_examples}(a), to compute the probability of reaching the goal safely at each state-action pair. Then, safe actions are identified by applying a user-defined threshold on the probability. However, the proposed model checker requires a discretization of the state space and a full transition model. Alternatively, Isele et al. \cite{isele2018safe} proposed the use of probabilistic prediction to identify potentially dangerous actions that would cause collision, but the safety guarantee may not be sufficiently strong if the prediction is not accurate. Prior knowledge \& constraints (e.g., lane changes are disallowed if they will lead to small time gaps) are also exploited \cite{mukadam2017tactical,mirchevska2018high,liu2019learning}.
For DIL, Zhang et al. \cite{zhang2016query} proposed SafeDAgger, in which a safety policy is learned to predict the error made by a primary policy without querying the reference policy. If the safety policy determines that it is unsafe to let the primary policy drive, the reference policy will take over. One drawback is that the quality of the learned policy may be limited by that of the reference policy.
\subsubsection {Combined Methods}
Various studies combine standard DRL/DIL with traditional rule-based methods to enhance safety. In contrast to the modified methods that are discussed above, combined methods don't modify the learning process of standard DRL/DIL. As presented in Fig. \ref{safety_tree_and_examples}(b), Chen et al. \cite{chen2019deep} proposed a framework in which DIL plans trajectories, while the rule-based tracking and safe set controller ensure safe control. Xiong et al. \cite{xiong2016combining} proposed the linear combination of the control output from DDPG, artificial potential field and path tracking modules. According to Shalev-Shwartz et al. \cite{shalev2016safe}, hard constraints should be injected outside the learning framework. They decompose the double-merge problem into a composition of a learnable DRL policy and a trajectory planning module with non-learnable hard constraints. The learning part enables driving comfort, while the hard constraints guarantee safety. 

\subsubsection {Hybrid Methods}
Hybrid methods integrate DRL/DIL into traditional heuristic search or POMDP planning methods. As presented in Fig. \ref{safety_tree_and_examples}(c), Bernhard et al. \cite{bernhard2018experience} integrated experiences in the form of pretrained Q-values into Hybrid A$^*$ as heuristics, thereby overcoming the statistical failure rate of DRL while still benefitting computationally from the learned policy. However, the experiments are limited to stationary environments. Pusse et al. \cite{pusse2019hybrid} presented a hybrid solution that combines DRL and approximate POMDP planning for collision-free autonomous navigation in simulated critical traffic scenarios, which benefits from advantages of both methods.
\vspace{-0.6cm}
\subsection {Interaction-aware DRL/DIL for AD}
\label{section:interaction}
Interaction is one of the intrinsic characteristics of traffic environments. An intelligent agent should reason beforehand about the behaviors of other traffic participants to passively react or actively adjust its own policy to cooperate or compete with other agents. 
This section reviews the interaction modeling methods and two groups of interaction-aware DRL/DIL methods for AD, as presented in Fig. \ref{interaction_tree_and_examples}. 
One group of methods focus on interactive environment encoding, while the other focus on interactive learning strategies.
\subsubsection {Interaction Modeling}
The simplest way to model interaction between multi-agents is to use a standard Markov decision process (MDP), where the other traffic participants are only treated as part of the environment \cite{chen2019crowd,huegle2019dynamic}. POMDP is another common interaction model \cite{bouton2019cooperation,qi2018intent,jiechuan2020graph}, where the agent has limited sensing capabilities. A Markov game (MG) is also used for modeling interaction scenarios. According to whether agents have the same importance, methods can be categorized into three groups: 1) equal importance \cite{mohseni2019interaction,hu2019interaction,fisac2019hierarchical}, 2) one vs. others \cite{li2017game}, and 3) proactive-passive pair \cite{ding2018game}. 
\subsubsection {Interactive Environment Encoding}
Interactive encoding of the environment is a popular research direction. As presented in Fig. \ref{interaction_tree_and_examples}, mainstream methods can be divided into two groups. One group of methods explicitly model other agents and utilize active reasoning about other agents in the algorithm workflow. POMDP is a common choice for these methods, where the intentions/cooperation levels of other agents are modeled as unobservable states that must be inferred. 
Qi et al. \cite{qi2018intent} proposed an intent-aware multi-agent planning framework, as presented in Fig. \ref{interaction_tree_and_examples}(a), which decouples intent prediction, high-level reasoning and low-level planning. The maintained belief regarding other agents' intents (objectives) was considered in the planning process. Bouton et al. \cite{bouton2019cooperation} proposed a similar method that maintains a belief regarding the cooperation levels (e.g., the willingness to yield to the ego vehicle) of other drivers.

The other group of methods focuses on utilizing special neural network architectures to capture the interplay between agents by their relation or interaction representations. These methods are usually agnostic regarding the intentions of other agents. Jiang et al. \cite{jiechuan2020graph} proposed graph convolutional reinforcement learning, in which the multi-agent environment is constructed as a graph. Agents are represented by nodes, and each node's corresponding neighbors are determined by distance or other metrics. Then, the latent features that are produced by graph convolutional layers are exploited to learn cooperation. Similarly, Huegle et al. \cite{huegle2019dynamicinteraction} built upon graph neural networks \cite{kipf17semi} and proposed the deep scene architecture for learning complex interaction-aware scene representations. Inspired by social pooling \cite{alahi2016social,gupta2018social} and attention models \cite{vemula2018social,hoshen2017vain}, Chen et al. \cite{chen2019crowd} proposed a socially attentive DRL method for interaction-aware robot navigation through a crowd. As illustrated in Fig. \ref{interaction_tree_and_examples}(b), they extracted pairwise features of interaction between the robot and each human and captured the interactions among humans via local maps. A self-attention mechanism was subsequently used to infer the relative importance of neighboring humans and aggregate interaction features. 
\subsubsection {Interactive Learning Strategy}
\begin{figure*}[h]
\centering
\includegraphics[width=0.85\textwidth]{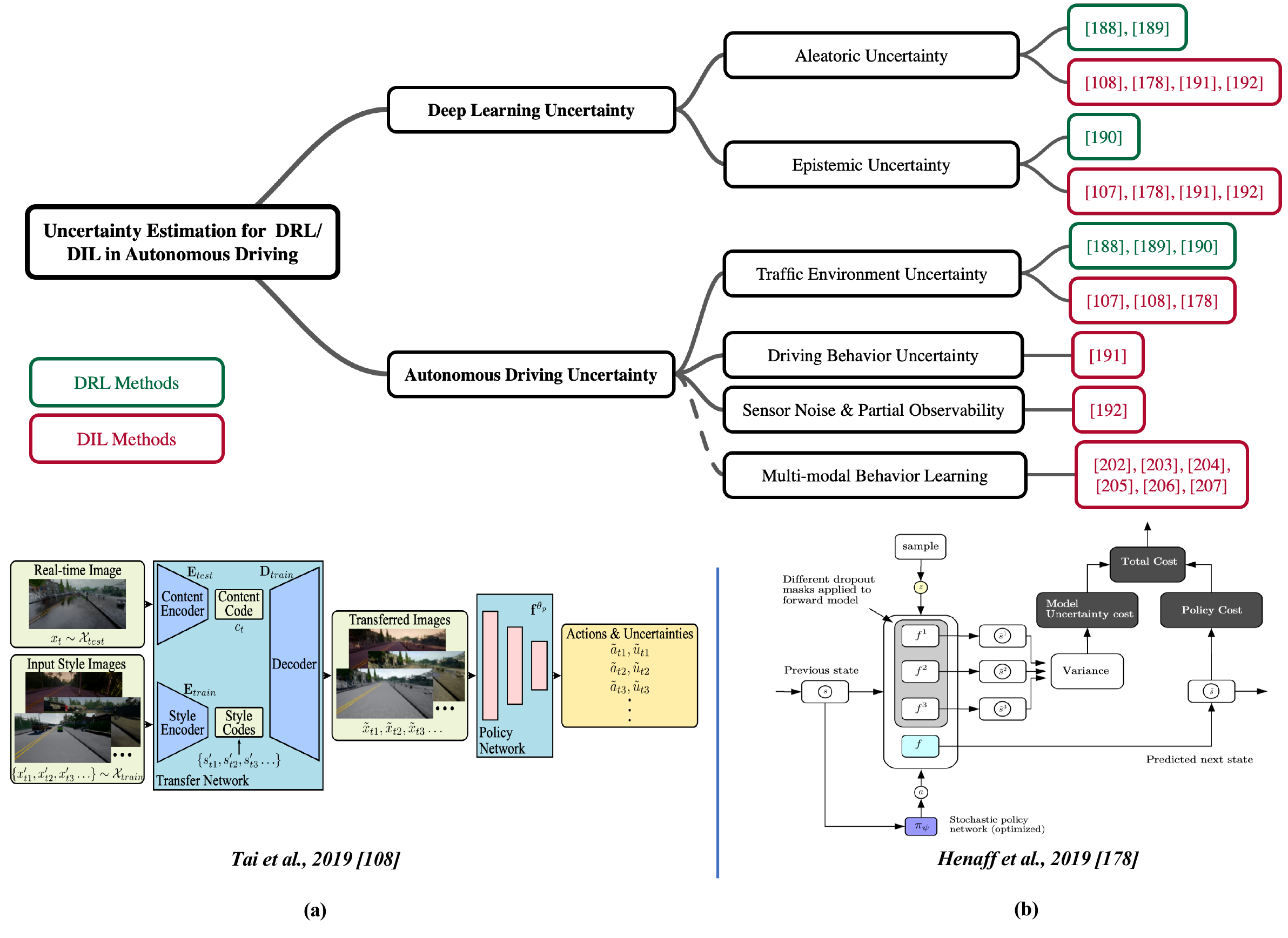}
\caption{A taxonomy of the literature on uncertainty-aware DRL/DIL models. (a) Tai et al. \cite{tai2019visual} addresses aleatoric uncertainty,  while (b) Henaff et al. \cite{henaff2019model} addresses both aleatoric and epistemic uncertainties.}
\label{uncertainty_tree_and_examples}
\vspace{-0.6cm} 
\end{figure*}
Various learning strategies have been used to learn interactive policies. Curriculum learning has been used to learn interactive policies \cite{hu2019interaction}\cite{mohseni2019interaction}, which can decouple complex problems into simpler problems. As presented in Fig. \ref{interaction_tree_and_examples}(c), Hu et al. \cite{hu2019interaction} proposed an interaction-aware decision making approach that leverages curriculum learning. First, a decentralized critic for each agent is learned to generate distinct behaviors, where the agent does not react to other agents and only learns how to execute rational actions to complete its own task. Second, a centralized critic is learned to enable agents to interact with each other to realize joint success and maintain smooth traffic. One limitation of these methods is that new models must be learned if the number of agents increases. 
Based on dynamic coordination graph (DCG) \cite{guestrin2002coordinated}, Chao et al. \cite{yu2019distributed} proposed a strategic learning solution for coordinating multiple autonomous vehicles in highways. DCG was utilized to explicitly model the continuously changing coordination dependencies among vehicles. Another group of interaction-aware methods use game theory\cite{leyton2008essential}. Game theory has already been applied in robotics tasks such as robust control \cite{papavassilopoulos1989robust,branch2007robust} and motion planning \cite{zhang1998motion,zhu2014game}. Recent years have also witnessed the increasing application of game theory in interaction-aware AD policy learning \cite{li2017game,fisac2019hierarchical,ding2018game}. Li et al. \cite{li2017game} proposed the combination of hierarchical reasoning game theory (i.e. ``level-$k$'' reasoning \cite{wilson1995on}) and reinforcement learning. Level-$k$ reasoning is used to model intelligent vehicles' interactions in traffic, while RL evolves these interactions in a time-extended scenario. 
Ding et al. \cite{ding2018game} introduced a proactive-passive game theoretical lane changing framework. The proactive vehicles learn to take actions to merge, while the passive vehicles learn to create merging space. Fisac et al. \cite{fisac2019hierarchical} proposed a novel game-theoretic real-time trajectory planning algorithm. The dynamic game is hierarchically decomposed into a long-horizon ``strategic'' game and a short-horizon ``tactical'' game. Furthermore, the long-horizon interaction game is solved to guide short-horizon planning, thereby implicitly extending the planning horizon and pushing the local trajectory optimization closer to global solutions. Apart from combining game theory and RL, solving an imitation learning problem under game-theoretic formalism is another approach. Sun et al. \cite{sun2018probabilistic} proposed an interactive probabilistic prediction approach that was based on hierarchical inverse reinforcement learning (HIRL). They modeled the problem from the perspective of a two-agent game by explicitly considering the responses of one agent to the other. However, some of the current game-theoretic interaction-aware methods are limited by their two-vehicle settings and simulation experiments \cite{ding2018game,fisac2019hierarchical,sun2018probabilistic}.
\subsection {Uncertainty-aware DRL/DIL for AD}
\label{section:uncertainty}
Before deployment of a learned model, it is important to determine what it does not understand and estimate the uncertainty of the decision making output. As presented in Fig. \ref{uncertainty_tree_and_examples}, this section reviews the uncertainty-aware DRL/DIL methods for AD from three aspects: 1) \textit{AD and deep learning uncertainty}, 2) \textit{uncertainty estimation methods}, and 3) \textit{multi-modal driving behavior learning}.
\subsubsection {Autonomous Driving and Deep Learning Uncertainty}
Autonomous driving has inherent uncertainty, while deep learning methods have deep learning uncertainty, and the two can intersect.
The AD uncertainty can be categorized as:
\begin{itemize}
\item \textbf{Traffic environment uncertainty} \cite{wang2019uncertainty,cui2019uncertainty,bernhard2019addressing,tai2019visual,kahn2017uncertainty,henaff2019model}. Stochastic and dynamic interactions among agents with distinct behaviors lead to intrinsic irreducible randomness and uncertainty in a traffic environment. 
\item \textbf{Driving behavior uncertainty} \cite{choi2018uncertainty}. Human driving behavior is multi-modal and stochastic (e.g., a driver can either make a left lane change or a right lane change when he comes up behind a van that is moving at a crawl).
\item \textbf{Partial observability and sensor noise uncertainty} \cite{lee2018safe}. In real-world scenarios, the AD agent usually has limited partial observability (e.g., due to occlusion), and there is noise in the sensor observation. 
\end{itemize}

Deriving from Bayesian deep learning approaches, Gal et al. \cite{gal2016uncertainty} categorized deep learning uncertainty as \textbf{aleatoric/data} and \textbf{epistemic/model uncertainties}. Aleatoric uncertainty results from incomplete knowledge about the environment (e.g., partial observability and measurement noise), which can't be reduced through access to more or even unlimited data but can be explicitly modeled. In contrast, epistemic uncertainty originates from an insufficient dataset and measures what our model doesn't know, which can be eliminated with sufficient training data.
 We refer readers to \cite{kendall2017uncertainties,gal2016uncertainty} for a deeper background on predictive uncertainty in deep neural networks. Although it is sometimes possible to use only aleatoric \cite{wang2019uncertainty,bernhard2019addressing,tai2019visual} or epistemic \cite{cui2019uncertainty,kahn2017uncertainty} uncertainty to develop a reasonable model, the ideal approach would be to combine these two uncertainty estimates \cite{lee2018safe,choi2018uncertainty,henaff2019model}. 
\subsubsection {Uncertainty Estimation Methods}
Aleatoric uncertainty is usually learned by using the heteroscedastic loss function \cite{kendall2017uncertainties}. The regression task and the loss are formulated as 
\begin{align}
& [\tilde{\mathbf{y}},\tilde{\sigma}] = \mathbf{f}^\theta(\mathbf{x}) \label{eqn_1}\\
& \mathcal{L}(\theta) = \frac{1}{2} \frac{{\parallel \mathbf{y} - \tilde{\mathbf{y}} \parallel}^2 }{\tilde{\sigma}^2} + \frac{1}{2}\log \tilde{\sigma}^2 \label{heteroscedastic_loss}
\end{align}
where $\mathbf{x}$ denotes the input data and $\mathbf{y}$ and $\mathbf{\tilde{y}}$ denote the regression ground truth and the prediction output, respectively. $\theta$ denotes the model parameters, and $\tilde{\sigma}$ is another output of the model, which represents the standard variance of data $\mathbf{x}$ (the aleatoric uncertainty). The loss function can be interpreted as penalizing a large prediction error when the uncertainty is small and relaxing constraints on the prediction error when the uncertainty is large. In practice, the network predicts the log variance $\log \tilde{\sigma}^2$ \cite{kendall2017uncertainties}.
Tai et al. \cite{tai2019visual} proposed an end-to-end real-to-sim visual navigation deployment pipeline, as illustrated in Fig. \ref{uncertainty_tree_and_examples}(a). An uncertainty-aware IL policy is trained with the heteroscedastic loss and outputs actions, along with associated uncertainties. A similar technique is proposed by Lee et al. \cite{lee2018safe}.

The epistemic uncertainty is usually estimated via two popular methods: Monte Carlo (MC)-dropout \cite{gal2015bayesian,gal2017deep} and ensembles \cite{dietterich2000ensemble,lakshminarayanan2017simple}. These methods are similar in the sense that both apply probabilistic reasoning on the network weights. The variance of the model output serves as an estimate of the model uncertainty. However, multiple stochastic forward passes through dropout sampling may be time-consuming, while ensemble methods have higher training and storage costs. 
Kahn et al. \cite{kahn2017uncertainty} proposed an uncertainty-aware RL method that utilizes MC-dropout and bootstrapping \cite{efron1993introduction}, where the confidence for a specified obstacle is updated iteratively. Guided by the uncertainty cost, the agent behaves more carefully in unfamiliar scenarios in the early training phase. As presented in Fig. \ref{uncertainty_tree_and_examples}(b), Henaff et al. \cite{henaff2019model} proposed training a driving policy by unrolling a learned dynamics model over multiple time steps while explicitly penalizing the original policy cost and an uncertainty cost that represents the divergence from the training dataset. Their method estimates both the aleatoric and epistemic uncertainties.

The uncertainty estimation methods that are presented above depend mainly on sampling. A novel uncertainty estimation method that utilizes a mixture density network (MDN) was proposed by Choi et al. \cite{choi2018uncertainty} for learning from complex and noisy human demonstrations. Since an MDN outputs the parameters for constructing a Gaussian mixture model (GMM), the total variance of the GMM can be calculated analytically, and the acquisition of uncertainty requires only a single forward pass. Distributional reinforcement learning \cite{bellmare2017distributional,dabney2018distributional,dabney2018implicit} offers another approach for modelling the uncertainty that is associated with actions. It models the RL return $R$ as a random variable that is subject to the probability distribution $Z(r|s,a)$ and the Q-value as the expected return $Q(s,a) = \mathbb{E}_{r\sim Z(r|s,a)}[r]$.
In the AD domain, Wang et al. \cite{wang2019uncertainty} applied distributional DDPG to an energy management strategy (EMS) problem as a case study to evaluate the effects of estimating the uncertainty that is associated with various actions at various states. Bernhard \cite{bernhard2019addressing} et al. presented a two-step approach for risk-sensitive behavior generation that combined offline distribution reinforcement learning with online risk assessment, which increased safety in intersection crossing scenarios.
\subsubsection {Multi-Modal Driving Behavior Learning}
\label{multi-modal}
The inherent uncertainty in driving behavior results in multi-modal demonstrations.
Many multi-modal imitation learning methods have been proposed. InfoGAIL \cite{li2017infogail} and Burn-InfoGAIL \cite{kuefler2017burn} infer latent/modal variables by maximizing the mutual information between latent variables and state-action pairs. VAE-GAIL \cite{wang2017robust} introduces a variational auto-encoder for inferring modal variables. However, due to the lack of labels in the demonstrations, these algorithms tend to distinguish latent labels without considering semantic information or the task context. Another direction focuses on labeled data in expert demonstrations. CGAIL \cite{merel2017learning} sends the modal labels directly to the generator and the discriminator. ACGAIL \cite{lin2018acgail} introduces an auxiliary classifier for reconstructing the modal information, where the classifier cooperates with the discriminator to provide the adversarial loss to the generator. Nevertheless, the above methods mainly leverage random sampling of latent labels from a known prior distribution to distinguish multiple modalities. The trained models rely on manually specified labels to output actions; hence, they cannot select modes adaptively according to environmental scenarios. Recently, Fei et al. \cite{fei20triple} proposed Triple-GAIL, which can learn adaptive skill selection and imitation jointly from expert demonstrations, and generated experiences by introducing an auxiliary skill selector. 

\section {Discussion}
\label{sec:challenge}
Although DRL and DIL attract significant amounts of interest in AD research, they remain far from ready for real-world applications, and challenges are faced at the architecture, task and algorithm levels. However, solutions remain largely underexplored. In this section, we discuss these challenges along with future investigation. DRL and DIL also have their own technical challenges; we refer readers to comprehensive discussions in \cite{arulkumaran2017deep,nguyen2018deep,osa2018algorithmic}.
\subsection{System architecture}
The success of modern AD systems depends on the meticulous design of architectures. The integration of DRL/DIL methods to collaborate with other modules and improve the performance of the system remains a substantial challenge. Studies have demonstrated various ways that DRL/DIL models could be integrated into an AD system. As illustrated Fig. \ref{fig:modes_all}, some studies propose new AD architectures, e.g., Modes 1\&2, where an entire pipeline from the input of the sensor/perception to the output of the vehicle's actuators is covered. However, the traditional modules of sequential planning are missing in these new architectures, and driving policy is addressed at the control level only. Hence, these AD systems could adapt to only simple tasks, such as road following, that require neither the guidance of goal points nor switching of driving behaviors. The extension of these architectures to accomplish more complicated AD tasks remains a substantial challenge. Other studies utilize the traditional AD architectures, e.g., Modes 3\&4, where DRL/DIL models are studied as substitutes for traditional modules to improve the performance in challenging scenarios. Mode 5 studies use both new and traditional architectures. Overall, the research effort until now has been focused more on exploring the potential of DRL/DIL in accomplishing AD tasks, whereas the design of the system architectures has yet to be intensively investigated. 
\subsection{Formulation of driving tasks}
Various DRL/DIL formulations have been established for accomplishing AD tasks. However, these formulations rely heavily on empirical designs. 
As reviewed in Section \ref{sec: task-driven}, the state space and input data are designed case by case, 
and ad-hoc reward functions are usually adopted with hand-tuned coefficients that balance the costs regarding safety, efficiency, comfort, and traffic rules, among other factors. 
Such designs are very brute-force approaches, which lack both theoretical proof and in-depth investigations.
Changing the designs or tuning the parameters could result in substantially different driving policies. 
However, in real-world deployment, more attention should be paid to the following questions: 
What design could realize the most optimal driving policy? Could such a design adapt to various scenes? How can the boundary conditions of designs be identified? To answer these questions, rigorous studies with comparative experiments are needed.

\subsection{Safe driving policy}
AD applications have high requirements on safety, and guaranteeing the safety of a DRL/DIL integrated AD system is of substantial importance. Compared to traditional rule-based methods, DNN has been widely acknowledged as having poor interpretability. Its ``black-box'' nature renders difficult the prediction of when the agent may fail to generate a safe policy. Deep models for real-world AD applications must address unseen or rarely seen scenarios, which is difficult for DL methods as they optimize objectives at the level of expectations over specified instances. To solve this problem, a general strategy is to combine traditional methods to ensure a DRL/DIL agent's functional safety. As reviewed in Fig. \ref{safety_tree_and_examples}, various methods have been proposed in the literature, where the problems are usually formulated as compositions of learned policies with hard constraints \cite{liu2019learning,mirchevska2018high,mukadam2017tactical}. However, balancing between the learned optimal policy and the safety guarantee by hard constraints is non-trivial and requires intensive investigation in the future.

\subsection{Interaction with traffic participants}
The capability of human-like interaction is required of self-driving agents for sharing the roads with other traffic participants. As reviewed in Section \ref{section:interaction}, interaction-aware DRL/DIL is a rising topic, but the following problems remain: First, current studies attempt to solve the problem from various perspectives, and the systematic studies are needed. Second, few interaction-aware DIL methods are available, while interaction-aware DRL methods are limited to simplified scenarios that involve only a few agents. The combination of interaction-aware trajectory prediction methods \cite{schmerling2018multimodal,li2019interaction,ma2019wasserstein} may be an open topic of potential value. Third, game theory and multi-agent reinforcement learning (\textbf{MARL}) \cite{nguyen2018deep} are highly correlated for interactive scenarios. MARL methods usually build on concepts of game theory (e.g., Markov games) to model the interaction process. Apart from DRL/DIL, methods are available for learning interactive policies through traditional game theory approaches, such as Nash equilibrium \cite{turnwald2016understanding}, level-$k$ reasoning \cite{tian2018adaptive} and game tree search \cite{isele2019interactive}. Exploiting POMDP planning to learn interactive polices is also a trend \cite{bai2015intention,hubmann18belief}. These methods have satisfactory interpretability but are limited to simplified or coarse discretizations of the agents' action space \cite{isele2019interactive,hubmann18belief}. Although the simplification reduces the computation burden, it tends to also lower the control precision. In the future, the combination of these methods and DRL/DIL may be promising. 

\subsection{Uncertainty of the environment}
Decision-making under uncertainty has been studied for decades \cite{holloway1979decision,kochenderfer2015decision}. Nevertheless, modeling the uncertainty in DRL/DIL formulations remains challenging, especially under complex uncertain traffic environments. Several problems have been identified in current research: First, most uncertainty-aware methods follow the style of deep learning predictive uncertainty \cite{gal2016uncertainty} without a deeper investigation. Is computing the predictive uncertainty of DNNs sufficient for AD tasks? Second, can the computed uncertainty be effectively utilized to realize a better decision making policy? Various methods incorporate the uncertainty cost into the global cost functions \cite{kahn2017uncertainty,henaff2019model}, while other methods utilize uncertainty to generate risk-sensitive behavior \cite{bernhard2019addressing}. Future efforts are needed to identify more promising applications. Third, human-driving behavior is uncertain, or multi-modal. However, DIL performs well for demonstrations from one
expert rather than multiple experts \cite{camacho1995behavioral}. A naive solution is to neglect the multi-modality and treat the demonstrations as if there is only one expert.
 The main side effect is that the model tends to learn an average policy rather than a multi-modal policy \cite{li2017infogail}. Thus, determining whether DRL/DIL learn effectively from noisy uncertain naturalistic driving data and generate multi-modal driving behavior according to various scenarios is meaningful. 

\subsection{Validation and benchmarks}
Validation and benchmarks are especially important for AD, but far from sufficient effort has been made regarding these aspects. First, comparison between DRL/DIL integrated architectures and traditional architectures is usually neglected in the literature, which is meaningful for identifying the quantitative performance gains and disadvantages of introducing DRL/DIL. Second, systematic comparison between DRL/DIL architectures is necessary. A technical barrier of the former two problems is the lack of a reasonable benchmark. High-fidelity simulators such as CARLA \cite{dosovitskiy2017carla} may provide a virtual platform on which various architectures can be deployed and evaluated. Third, exhaustive validation of trained policies before deployment is of vital importance. However, validation is challenging. Real-world testing on vehicles has high costs in terms of time, finances and human labor and could be dangerous. Empirical validation through simulation can reduce the amount of required field testing and can be used as a first step for performance and safety evaluation. However, verification through simulation only ensures the performance in a statistical sense. Even small variations between the simulators and the real scenario can have drastic effects on the system behavior.
Future studies are needed to identify practical, effective, low-risk and economical validation methods.

\section {Conclusions}
\label{sec:conclusion}
In this study, a comprehensive survey is presented that focuses on autonomous driving policy learning using DRL/DIL, which is addressed simultaneously from the system, task-driven and problem-driven perspectives. The study is conducted at three levels: First, a taxonomy of the literature studies is presented from the system perspective, from which five modes of integration of DRL/DIL models into an AD architecture are identified. Second, the formulations of DRL/DIL models for accomplishing specified AD tasks are comprehensively reviewed, where various designs on the model state and action spaces and the reinforcement learning rewards are covered. Finally, an in-depth review is presented on how the critical issues of AD applications regarding driving safety, interaction with other traffic participants and uncertainty of the environment are addressed by the DRL/DIL models. The major findings are listed below, from which potential topics for future investigation are identified.
\begin{itemize}
\item DRL/DIL attract significant amounts of interest in AD research. However, literature studies in this scope have focused more on exploring the potential of DRL/DIL in accomplishing AD tasks, whereas the design of the system architectures remains to be intensively investigated.
\item Many DRL/DIL models have been formulated for accomplishing AD tasks. However, these formulations rely heavily on empirical designs, which are brute-force approaches and lack both theoretical proof and in-depth investigations. In the real-world deployment of such models, substantial challenges in terms of stability and robustness may be encountered.
\item Driving safety, which is the main issue in AD applications, has received the most attention in the literature. However, the studies on interaction with other traffic participants and the uncertainty of the environment remain highly preliminary, in which the problems have been addressed from divergent perspectives, and have not been conducted systematically.
\end{itemize}
\footnotesize
\bibliographystyle{IEEEtran}
\bibliography{ref.bib}
\begin{IEEEbiography}[\vspace{-10mm}{\includegraphics[width=1in,height=1in,clip,keepaspectratio]{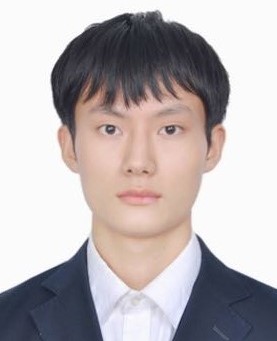}}]{Zeyu Zhu}
	received B.S. degree in computer science from Peking University, Beijing, China, in 2019, where he is currently pursuing the Ph.D. degree with the Key Laboratory of Machine Perception (MOE), Peking University.
	His research interests include intelligent vehicles, reinforcement learning, and machine learning.
\end{IEEEbiography}
\vspace{-18mm}
\begin{IEEEbiography}
	[{\includegraphics[width=1in,height=1in,clip,keepaspectratio]{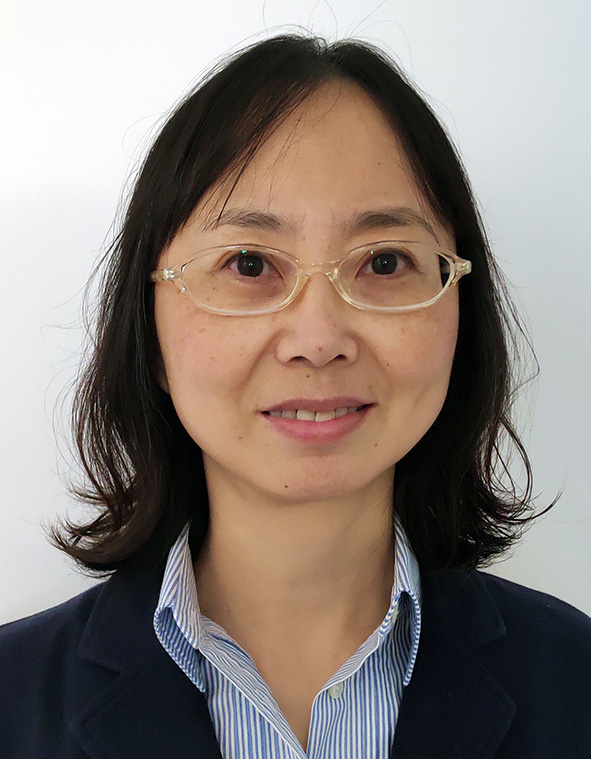}}]{Huijing Zhao}
received B.S. degree in computer science from Peking University in 1991. She obtained M.E. degree in 1996 and Ph.D. degree in 1999 in civil engineering from the University of Tokyo, Japan. From 1999 to 2007, she was a postdoctoral researcher and visiting associate professor at the Center for Space Information Science, University of Tokyo. In 2007, she joined Peking University as a tenure-track professor at the School of Electronics Engineering and Computer Science. She became an associate professor with tenure on 2013 and was promoted to full professor on 2020. She has research interest in several areas in connection with intelligent vehicle and mobile robot, such as machine perception, behavior learning and motion planning, and she has special interests on the studies through real world data collection.
\end{IEEEbiography}

\end{document}